\documentclass[final]{cvpr}

\usepackage[numbers,sort,compress]{natbib}
\usepackage{url}            
\usepackage{booktabs}       
\usepackage{amsfonts}       
\usepackage{nicefrac}       
\usepackage{microtype}      
\usepackage{graphicx}
\usepackage{amsmath}
\usepackage{amssymb}
\usepackage{bm}
\usepackage{makecell}
\usepackage{comment}
\usepackage{wrapfig}
\usepackage{subfig}
\usepackage{multirow}

\def\cvprPaperID{8866} 
\def\confYear{CVPR 2021}

\newcommand{\mbf}[1] {\mathbf{#1}}

\newcommand{\xx}{\mathbf{x}}
\newcommand{\vv}{\mathbf{v}}

\newcommand{\zz}{\mathbf{z}}
\newcommand{\II}{\mathbf{I}}
\newcommand{\uu}{\mathbf{u}}

\newcommand{\cut}[1]{}
\newcommand{\keypoint}[1]{\noindent\textbf{#1}\quad}
\newcommand{\doublecheck}[1]{#1}

\begin{document}

\title{NewtonianVAE: Proportional Control and Goal Identification from Pixels via Physical Latent Spaces}

\author{%
  Miguel Jaques \\
  University of Edinburgh\\
  Edinburgh, UK \\
  {\tt\small m.a.m.jaques@sms.ed.ac.uk}
    \and
    Michael Burke \\
  Monash University\\
  Melbourne, AU \\
  {\tt\small michael.burke1@monash.edu}
  \and
  Timothy Hospedales \\
  University of Edinburgh \\
  Edinburgh, UK \\
  {\tt\small t.hospedales@ed.ac.uk}
}

\maketitle

\begin{abstract}
  Learning low-dimensional latent state space dynamics models has proven powerful for enabling vision-based planning and learning for control. We introduce a latent dynamics learning framework that is uniquely designed to induce \emph{proportional} controlability in the latent space, thus enabling the use of simple and well-known PID controllers. We show that our learned dynamics model enables proportional control from pixels, dramatically simplifies and accelerates behavioural cloning of vision-based controllers, and provides interpretable goal discovery when applied to imitation learning of switching controllers from demonstration. Notably, such proportional controlability also allows for robust path following from visual demonstrations using Dynamic Movement Primitives in the learned latent space.
\end{abstract}

\vspace{-5mm}

\section{Introduction}



Vision-based control is highly desirable across numerous industrial applications, both in robotics and process control. At present, much practical vision-based control relies on supervised learning to build bespoke perception modules, prior to downstream dynamics modelling and controller design. This can be expensive and time consuming, and as a result there is growing interest in developing model-based approaches for direct vision-based control.

Model-based approaches for visual control tend to learn latent dynamics models that are subsequently used within suitable planning or model predictive control (MPC) frameworks, or to train policies for later use. We argue that this decoupling of dynamics and control is computationally expensive and often unnecessary. Instead we learn a structured latent dynamical model that directly allows for simple proportional control to be applied. 
Proportional-Integral-Derivative (PID) feedback control produces commands
that are proportional to an error or cost term between current system state $\xx$ and a (potentially dynamic)  target state $\xx^{goal}$:\vspace{-2mm}
\begin{eqnarray}
    \uu_t = K_p\,(\xx_t^{goal}-\xx_t) + K_i \sum_{t'}(\xx_{t'}^{goal}-\xx_{t'}) + \nonumber \\ 
    K_d\, \frac{\xx_t-\xx_{t-1}}{\Delta t} 
    \label{eq:teaser}
\end{eqnarray}
Gain terms $(K_p, K_i, K_d)$ shape the controller response to errors. 
PID control is ubiquitous in industry, and broadly applicable across numerous domains, providing a simple and reliable off-the-shelf mechanism for stabilising systems. PID control is also the basis of a wide range of more powerful control strategies, including the more flexible dynamic movement primitives \cite{schaal2006dynamic,ijspeert2013dynamical} that augment PD control laws with a forcing function for trajectory following. \doublecheck{Essentially we learn the state encoding $\xx(I)$ from images $I$ for which robots can be trivially controlled from pixels according to Eq~\ref{eq:teaser}.}

We structure latent dynamics so that that PID control can be applied to move between latent states, to remove the requirement for complex planning or reinforcement learning strategies.  Moreover, we show that imitation learning from demonstrations becomes a simple goal inference problem under a proportional control model in this latent space, and can even be extended to sequential tasks comprising multiple sub-goals.

Imitation learning from high dimensional visual data is particularly challenging \cite{Bagnell-2015-5921}. 
Behaviour cloning, which seeks to reproduce demonstrations, is particularly vulnerable to generalisation failures for high dimensional visual inputs,
while inverse reinforcement learning (IRL) \cite{ng2000algorithms} strategies are hard to train and extremely sample inefficient. By learning a structured dynamics model, we allow for more robust control in the presence of noise and simplify the inverse reward inference process.
In summary, the primary contributions of this work are:

\noindent\textbf{Embedding for proportional controllability} {} We induce a latent space where taking an action in the direction between the current position and some target position, $\uu \propto \xx^{target}-\xx$, moves the system towards the target position. Uniquely, this enables simple proportional control from pixels.

\noindent\textbf{Imitation learning using latent switching proportional control laws} {} We leverage the properties of this embedding to frame imitation learning as a goal inference problem under a switching proportional control law model in the structured latent space for sequential goal reaching problems. This enables one-shot interpretable imitation learning of switching controllers from high-dimensional pixel observations.

\noindent\textbf{Imitation learning using dynamic movement primitives (DMPs)} {} We also leverage the properties of our embedding to fit dynamic movement primitives in the structured latent space for trajectory tracking problems. This enables one-shot imitation learning of trajectory following controllers from pixels.

Results show that embedding for proportional controllability produces more interpretable latent spaces, allows for the use of simple and efficient  controllers that cannot be applied with less structured latent dynamical models, and enables one-shot learning of control and interpretable goal identification in sequential multi-task imitation learning settings.

\section{Related Work}

This paper takes a model-based approach to visual control, using variational autoencoding  (VAE) \cite{kingma2013auto}. Latent dynamical systems modelling using autoencoding is widely used \cite{lesort2018state}, and has been proposed for Bayesian filtering \cite{Fraccaro,Krishnan,Karl}, and as inverse graphics for improved video prediction and vision-based control \cite{Jaques2020}. \citet{ha2018world} train a latent dynamics model using a variational recurrent neural network (VRNN) in the latent space of a VAE, and then learn a controller that acts in this space using a known reward model. \citet{hafner2018learning} extend this approach to allow planning from pixels. Unfortunately, because these approaches decouple dynamics modelling and control, they place an unnecessary computational burden on control, either requiring sampling-based planning or further RL policy optimisation. We argue that this burden can be alleviated by imposing additional structure on the latent space such that proportional control becomes feasible.

In doing so, we build on the control hypothesis advocated by \citet{full1999templates}, which seeks to model complex phenonoma and systems through simple template models and controllers, using anchor networks to abstract the complexity away from control. This also simplifies the challenges of imitation learning, allowing for  sequential task composition \cite{burridge1999sequential}.

The addition of structural inductive biases into neural models has become increasingly important for generalisation. Injecting knowledge of known physical equations \cite{LeGuen,Jaques2020} has been shown to improve dynamics modelling, while the inclusion of structured transition matrices was essential to learn Koopman operators \cite{Abraham2017Model-BasedOperators} that model dynamical systems with compositional properties \cite{Li2020Learning}. Here, a block-wise structure with shared blocks was used to learn transition dynamics, which highlighted the importance of added structure in linear state space models, but this was not applied to visual settings. Models like embed to control (E2C) \cite{Watter} or deep variational Bayes filters (DVBF) \cite{Karl} recover structured conditionally linear latent spaces which can be used for control, but, as will be demonstrated later, are still unsuitable for direct proportional control. PVEs \cite{Jonschkowski2017PVEs:Representations} learn an explicit positional representation, but do so by minimizing a combination of several heuristic loss functions. Since these models do not use a decoder, it is not possible to visually inspect the learned representations in image space. 

NewtonianVAE not only provides latent space interpretability, but also simplifies imitation learning. Inverse reinforcement learning (IRL) strategies for imitation learning typically struggle to learn from high dimensional observation traces as they tend to be based on the principle of feature counting and observation frequency matching \cite{ng2000algorithms}, as in maximum entropy IRL \cite{ziebart2008maximum}. Maximum entropy IRL has been extended to use a deep neural network feature extractor \cite{ziebart2008maximum}, but this  is highly vulnerable to overfitting and has extensive data requirements. Recent adversarial IRL approaches \cite{fu2017learning,ghasemipour2019divergence,ho2016generative}  avoid the challenge of learning a global reward function by training policies directly, but these have yet to be successfully scaled to high dimensional problems. As a result, most imitation learning approaches tend to assume access to low dimensional states, avoiding the challenge of learning from pixels.

Behaviour cloning approaches using dynamic movement primitives (DMP) \cite{schaal2006dynamic,ijspeert2013dynamical} have proven particularly powerful for trajectory following control, but are typically applied to low-dimensional proprioceptive states directly as they require proportionally controllable state spaces. Deep DMPs \cite{pervez2017learning} learn visually task parametrised DMPs, but the DMP itself still requires low dimensional state measurements. \citet{chen2016dynamic} propose VAE-DMPs, which impose DMP dynamics in the latent space of a variational auto-encoder, allowing for direct imitation learning. In contrast, this work learns dynamics models independently of tasks, which allows for more flexible downstream applications, including DMP fitting for trajectory following and switching multi-goal imitation learning from pixels (unlike \citet{chen2016dynamic}, which use proprioception observations).

Standard imitation learning  learning strategies can fail in multi-goal settings or on more complex tasks. In order to address this, many approaches frame the problem of imitation learning from these lower level states as one of skill or options \cite{sutton1999between, konidaris2009skill} learning using switching state space models. These switching models include linear dynamical attractor systems \cite{Dixon04}, conditionally linear Gaussian models \cite{Chiappa10,levine2014learning}, Bayesian non-parametrics \cite{Niekum11,ranchod2015nonparametric}, and neural variational models \cite{Kipf2019CompILE:Execution}. \citet{Kipf2019CompILE:Execution} learn task segmentations to infer compositional policies, but the model uses environment states directly instead of images.  \citet{burke2019hybrid,burke2019explanation} use a switching controller formulation for control law identification from image, proprioceptive state and control action observations. This work applies a similar strategy for goal inference, but, unlike the approaches above, makes use of a learned latent state representation and does not require proprioceptive or low level state information.

Despite this reliance on proprioceptive state information, there is a growing interest in direct visual imitation learning and control. \citet{nair2018visual} train a variational autoencoder (VAE) on image observations of an environment, and subsequently sample from this latent space in order to train goal-conditioned policies that can be used to move between different goal states. In contrast, we propose a latent dynamics model that allows for latent proportional controllability and eliminates the need to train a policy to move between goal states. 

In addition to the works discussed above, a research area in the unsupervised learning literature of particular interest is that of learning physically plausible representations (from video) by enforcing temporal evolution according to explicit or implicit physical dynamics \cite{Belbute-Peres2018End-to-EndControl,Greydanus2019HamiltonianNetworks,Jaques2020,Toth2020HamiltonianNetworks}. Though promising, these approaches have only been applied to very simple toy environments where dynamics are well known, and are still to be scaled up to real world scenes.

\section{Variational models for visual control}

In order to learn a compact latent representation of videos that can be used for planning and control we use the variational autoencoder framework (VAE) \cite{kingma2013auto,rezende2014stochastic} and its recurrent formulation (VRNN), \cite{Chung}. In this section we briefly present a general formulation of the VRNN, of which many recent models are particular cases or variations \cite{Watter,Karl,Fraccaro,Krishnan,hafner2018learning}. For derivation details please refer to \cite{Chung}.
\begin{figure}[!t]
  \centering
    \includegraphics[width=0.38\linewidth]{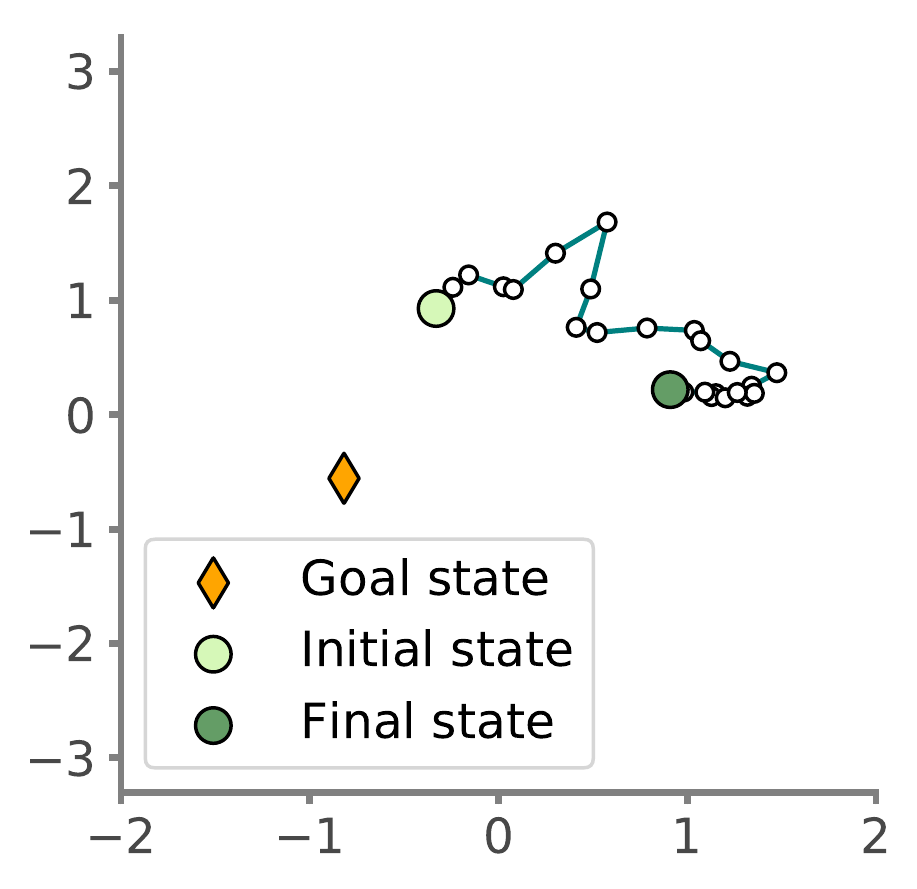}
    \hspace{5mm}
    \includegraphics[width=0.38\linewidth]{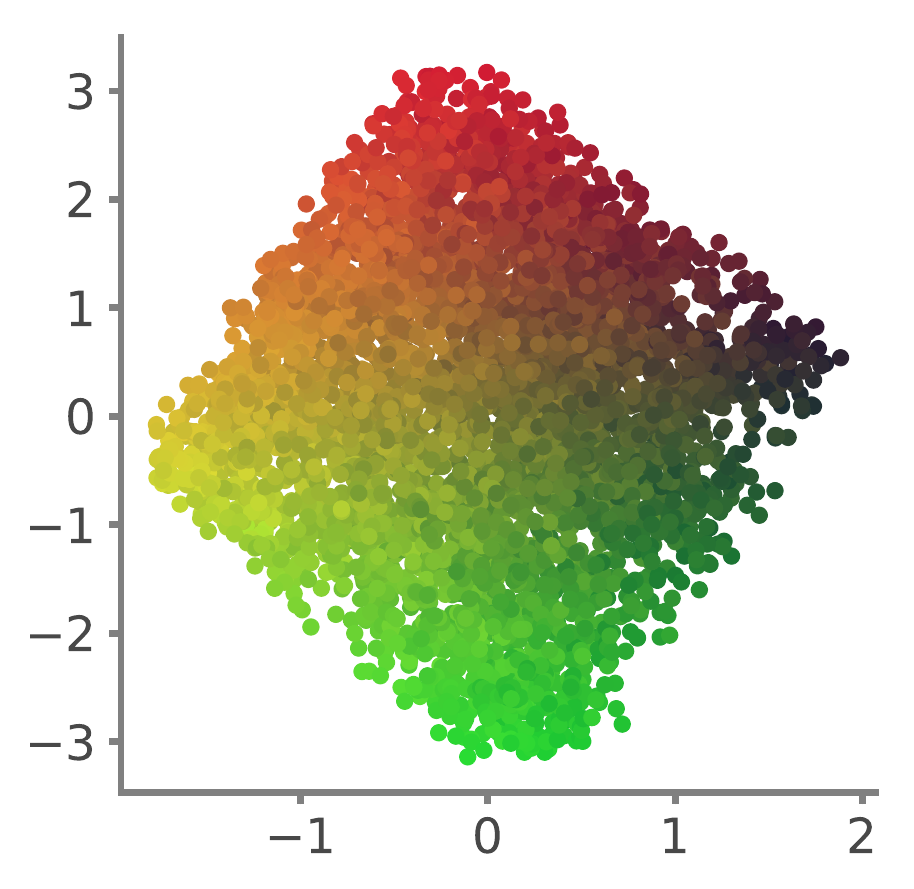}
    \vspace{-2mm}
    \caption{Trajectory of a point mass actuated using $\uu_t \propto (\xx^{goal} - \xx_t)$ (left) in the latent space learned by an E2C model (right).}
    \label{fig:motivation}
    \vspace{-4mm}
\end{figure}

Given a sequence of $T$ images, $\II_{1:T}$, and actuations $\uu_{1:T} \in \mathbb{R}^{d_u}$ and the corresponding latent representations, $\zz_{1:T} \in \mathbb{R}^{d_z}$, the marginal image likelihood is given by:
\begin{equation}
    \hspace{-2mm}p(\II_{1:T}|\uu_{1:T}) = \hspace{-1mm}\int \hspace{-1mm} p(\II_{1:T}|\zz_{1:T},\uu_{1:T}) p(\zz_{1:T}|\uu_{1:T}) \text{d}\zz_{1:T}
    \label{eq:marginal_ll}
\end{equation}
where we factorize the terms above as:
\begin{align*}
    p(\II_{1:T}|\zz_{1:T}, \uu_{1:T}) &= \prod p(\II_t| \zz_t) \\
    p(\zz_{1:T}|\uu_{1:T}) &= \prod p(\zz_t|\zz_{t-1}, \uu_{t-1}),
\end{align*}
with an approximate positerior given by:
\begin{equation}
   q(\zz_{1:T}|\II_{1:T}) = \prod q(\zz_t|\II_t, \zz_{t-1}, \uu_{t-1}).
\end{equation}
The model components are trained jointly by maximizing the lower bound on \eqref{eq:marginal_ll}:
\begin{align}
    \mathcal{L} = & \sum_t \mathbb{E}_{q(\zz_t |\II_t, \zz_{t-1}, \uu_{t-1})} \big[ p(\II_t|\zz_t)  + \nonumber \\ &+ \text{KL}\left(q(\zz_{t+1}|\II_{t+1}, \zz_t, \uu_t) \|  p(\zz_{t+1}|\zz_t, \uu_t) \big)  \right], \label{eq:elbo}
\end{align}
via the reparametrization trick, by drawing samples from the posterior distributions, $q(\zz_t |\II_t, \zz_{t-1}, \uu_{t-1})$. Under this framework, the various desired inductive biases are usually built into the structure of the transition prior $p(\zz_{t+1}|\zz_t, \uu_t)$. In this work we will build on the formulation that uses a linear dynamical system as latent dynamics:
\begin{equation}
    p(\zz_{t+1}|\zz_t, \uu_t)= A(\zz_t)\cdot\zz_t + B(\zz_t)\cdot\uu_t + \mbf{c}(\zz_t) \label{eq:lds}
\end{equation}
which has been studied extensively in the context of deep probabilistic models \cite{Linderman2017BayesianSystems,Fraccaro,Karl,Krishnan,Becker-Ehmck2019SwitchingFiltering}.

\section{Newtonian Variational Autoencoder}

\keypoint{Motivation} To motivate our model, we begin by examining the properties of an existing latent variable model used for control. We train an E2C model \cite{Watter}, since it applies a locally linear latent transition as in \eqref{eq:lds} and is highly representative of properties obtained in these types of model. We use a simple point mass system that can move in the $[x,y]$ plane and train the model on random transitions in image space (more details in the experiments section).  Since the environment is 2D with 2D controls, we use a 4D latent space (2 dimensions for position and 2 for velocity). Our goal is to explore how the E2C model behaves when a basic proportional control law $\uu_t \propto (\xx^{goal} - \xx_t)$ is applied, where $\xx$ is the latent system configuration.

 An immediate problem is that even though the latent coordinates corresponding to position are correctly learned (Fig. \ref{fig:motivation}(right)), it is necessary to plot \textit{every coordinate pair} and their correlation with ground truth positions in order to visually determine which 2 coordinates correspond to the position $\xx$. Having determined such $\xx$, we can use a random target position $\xx^{goal}$ and see if successively applying an action $\uu_t \propto (\xx^{goal} - \xx_t)$ will guide the system towards $\xx^{goal}$ (which we term \textit{proportional controllability}). Note that PID control is trivially achievable given a P-controllable system, so we focus on P-control for simplicity of exposition, without loss of generality. Fig. \ref{fig:motivation}(left) shows that this simple control law fails to guide the system towards the goal state, even though the latent space is seemingly well structured. These problems are present in existing variational models for controllable systems, including E2C \cite{Watter}, DVBF \cite{Karl} and the Kalman VAE \cite{Fraccaro}.


To avoid the need for ground truth data and visual inspection, we  construct a model that explicitly treats position and velocity as separate latent variables $\xx$ and $\vv$. To ensure correct behaviour under a proportional control law\footnote{For further analysis of the convergence and stability of PID controllers, see \cite{duc2006stability,dorf2011modern}.} the change in position and velocity should be directly related to the force applied. I.e. given an external action $\uu$ representing the force (=acceleration) acting on a system, $\xx$ and $\vv$ should follow Newton's second law, $ d^2\xx/dt^2 = \mbf{F}/m$. Although this might seem like a trivial statement from a physical standpoint, this type of behaviour is not built into existing neural models, where the relationship between action and latent states can be arbitrary. This arbitrary relationship in turn complicates  control, and it becomes necessary to learn downstream controllers or policies to compensate for these dynamics while meeting a control objective.


We make one additional observation: in many cases the external action $\uu$ is  applied along disentangled dimensions of the system. For example, for a 2-arm robot, actions correspond to torques on the angles of each arm relative to its origin\footnote{The example also applies more generally to any robot actuated with torques along its joints.}. These action dimensions correspond to the polar coordinates $[\theta_1, \theta_2]$, which are the ideal disentangled coordinates to describe such a robot. We use this fact to formulate a model that not only provides an interpretable and P-controllable latent space, but also the correct disentanglement by construction.

\keypoint{Formulation} We now formulate a model satisfying the above desiderata. For an actuated rigid body systems with $D$ degrees of freedom, we model the system configuration (positions or angles) by a set of coordinates $\xx \in \mathbb{R}^D$ with double integrator dynamics, inspired by Newton's equations of motion:
\begin{equation}
   \frac{d\xx}{dt} = \vv \, , \, \frac{d\vv}{dt} = A(\xx, \vv) \cdot \xx + B(\xx, \vv)\cdot \vv + C(\xx, \vv)\cdot \uu \label{eq:act_motion}
\end{equation}

To build a discrete form of \eqref{eq:act_motion} into a VAE formulation, we use the instantaneous system configuration (or position) $\xx$ as the stochastic variable that is inferred by the approximate posterior, $\xx_t \sim q(\xx_t | \II_t)$, with velocity a deterministic variable that is simply the finite difference of positions, $\vv_t = (\xx_t - \xx_{t-1})/ \Delta t$. The generative model is now given by 
\begin{align}
p(\II_{1:T}|\xx_{1:T}, \uu_{1:T}) &= \prod p(\II_t| \xx_t) \\
p(\xx_{1:T}|\uu_{1:T}) &= \prod p(\xx_t|\xx_{t-1}, \uu_{t-1}; \vv_t)
\end{align}
where the transition prior is:
\begin{align}
&p(\xx_t|\xx_{t-1}, \uu_{t-1}; \vv_t) = \mathcal{N}(\xx_t|\xx_{t-1} + \Delta t \cdot \vv_t, \; \sigma^2) \\
&\vv_t = \vv_{t-1} + \Delta t \cdot  (A \xx_{t-1} + B \vv_{t-1} + C \uu_{t-1}) \label{eq:nvae_trans}
\end{align}
with $[A,\log(-B),\log C]=\text{diag}(f(\xx_t, \vv_t, \uu_t))$, where $f$ is a neural network with linear output activation. Using diagonal transition matrices encourages correct coordinate relations between $\uu$, $\xx$ and $\vv$, since linear combinations of dimensions are eliminated\cut{\footnote{These can still exist via the network producing A,B,C, but this dependency does not invalidate our hypothesis, as shown empirically in experiments.}}. In order to obtain the correct directional relation between $\uu$ and $\xx$, required for interpretable controllability, we set $C$ to be strictly positive (in addition to diagonal). $B$ is strictly negative to provide a correct interpretation of the term in $\vv$ as friction, which aids trajectory stability. During inference, $\vv_t$ is computed as $\vv_t = (\xx_t - \xx_{t-1})/\Delta t$, 
with $\xx_t \sim q(\xx_t | \II_t)$ and $\xx_{t-1} \sim q(\xx_{t-1} | \II_{t-1})$.
This inference model provides a principled way to infer velocities from consecutive positions, similarly to \cite{Jonschkowski2017PVEs:Representations}.  We use Gaussian $p(\II_t|\xx_t)$ and $q(\xx_t|\II_t)$ parametrized by a neural network throughout.

We train all model components using the following ELBO (full derivation in Appendix A):
\begin{align}
    \mathcal{L} &= \mathbb{E}_{q(\xx_t | \II_t)q(\xx_{t-1} | \II_{t-1})} [ \mathbb{E}_{p(\xx_{t+1}|\xx_t, \uu_t; \vv_t)}  p(\II_{t+1}|\xx_{t+1}) + \nonumber  \\
    & + \text{KL}\left(q(\xx_{t+1}|\II_{t+1}) \|  p(\xx_{t+1}|\xx_t, \uu_t; \vv_t) \right) ]
    \label{eq:nvae_elbo}
\end{align}
A crucial component of this ELBO is performing  future- rather than current-step reconstruction through the generative process (first term above). This 
 is known to encourage the use of the  transition prior when learning the latent representation \cite{Watter,Karl,hafner2018learning}. 

\keypoint{Further considerations} Another key difference between a simple LDS and our Newtonian model is the fact that we consider velocity to be a deterministic latent variable that is uniquely determined by the stochastic positions.  In contrast, independent inference through $\zz$ means that position and velocity might not have the direct relation that is present in the physical world (velocity as the derivative of position). Both of these contribute to a lack of physical plausability, in the Newtonian sense, in existing models. Though technically our transition prior is a special case of the LDS \eqref{eq:lds}, these added structural constraints are crucial in order to induce a Newtonian latent space that directly allows for PID control of latent image states.

\section{Efficient Imitiation with P-Control}

A key benefit of the Newtonian latent space is that it dramatically simplifies image-based imitation learning.
Given a visual demonstration sequence $D_{\II} = \{(\II_1,\uu_1) , ..., (\II_T,\uu_T)\}$, we encode the frames using the inference network $q(\xx|\II)$ described above in order to produce demonstrations in latent space, $D_{\xx} = \{(\xx_1, \uu_1) , ..., (\xx_T, \uu_T)\}$. 

\subsection{Learning Vision-Driven Switching P-Control}
We can fit a switching P-controller\footnote{We use a P-controller instead of a PID-controller for simplicity of exposition and without loss of generality.} to a set of demonstration sequences in latent space using a Mixture Density Network (MDN), where the action likelihood given a state is a mixture of $N$  proportional controllers:
\vspace{-1.5mm}
\begin{equation}
    \hspace{-2mm} P(\uu_t|\xx_t) = \sum_{n=1}^N \pi_n(\xx_t) \, \mathcal{N}\left(\uu_t|K_n(\xx_n^{goal}-\xx_t),\, \sigma^2_n\right) \label{eq:mdn}
    \vspace{-1.5mm}
\end{equation}
where $K_n$, $\xx_n^{goal}$ and $\sigma^2_n$, $\forall n\in {1..N}$, are learnable parameters, and $\bm{\pi}(\zz)$ is a parametric function like a neural network. Intuitively, fitting this MDN to the latent demonstrations splits the demonstrations into regions where a specific proportional controller would correctly fit that part of the trajectory. If the latent space is P-controllable (such as the one produced by the NewtonianVAE), the vectors $\xx_n^{goal}$ will correspond to the intermediate goals or bottleneck states in the demonstration sequence. As an added benefit, we can pass the learned goals through NewtonianVAE's decoder in order to obtain their visual representation, providing an interpretable control policy.

\keypoint{Learning a finite-state machine} 
Having identified the latent vectors corresponding to the goals, we determine the order in which they must be reached by analysing their visits during the demonstrations, directly extracting initiation sets and termination conditions. This produces a simple finite-state machine (FSM) that determines goal state transitions. The FSM and extracted P-controllers can then be used to reproduce demonstrated behaviours by driving the robot to each goal in succession, but could also be used within an options framework \cite{sutton1999between} for reinforcement learning. 


\begin{figure}[!t]
    \centering
    \includegraphics[width=0.47\textwidth]{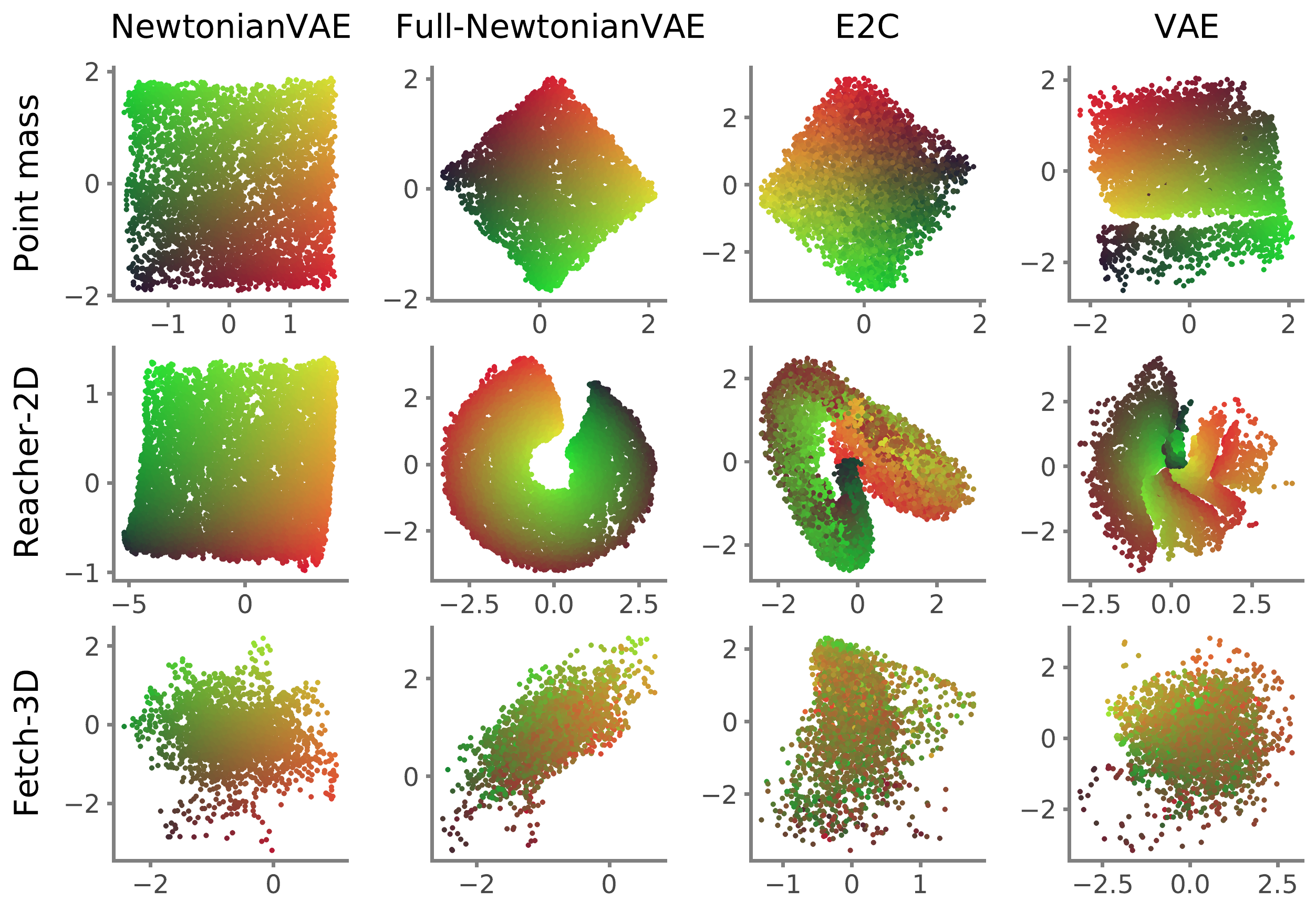}
    \vspace{-4mm}
    \caption{Latent spaces of various models in the point mass, reacher-2D and fetch-3D environments. Each dot corresponds to the latent representation of a test frame, and the red-to-green color coding encodes the true 2D position/angle values. For E2C \cite{Watter}, we plot the two latent dimensions that best correlated with the true positions. Since the configuration space of the fetch-3D env is 4D, we visualize only the first two coordinates. Only for our NewtonianVAE does latent space (position) and true space (color) correlate perfectly.}
    \label{fig:latent_spaces}
    \vspace{-3mm}
\end{figure}

\subsection{Learning Visual Path Following with DMPs}

It is clear that the latent space of a NewtonianVAE can be used for switching goal-based imitation learning, but proportionality is also a precursor for trajectory following using DMPs. A DMP \cite{ijspeert2013dynamical} is a proportional-derivative controller with a learned forcing function
\begin{equation}
    \tau\mathbf{\ddot{x}} = \alpha\left(\beta\left(\mathbf{x}^\text{goal}-\mathbf{x} \right) - \mathbf{\dot{x}}\right) + \mathbf{f}. \label{eq:dmp}
\end{equation}
Here, $\tau$ is a time scaling constant, and $\alpha,\beta$ are proportional control gain terms. The forcing function
\begin{equation}
\mathbf{f}_t = \sum_{i=1}^N \frac{\Phi(t)\mathbf{w}_i}{\sum_{i=1}^N \Phi(t)}(\mathbf{x}-\mathbf{x}^\text{goal})
\end{equation}
captures trajectory dynamics, using a weighted linear combination of radial basis functions, 
$
\Phi(t) = \text{exp}(-\frac{1}{\sigma_i^2}(y-c_i)^2)$,
with centres $c_i$ and variances $\sigma_i^2$. 

The canonical system $\dot{y} = -\alpha_y y$ gently decays over time, smoothly modulating the forcing function until reaching an end goal, $\mathbf{x}^\text{goal}$. Basis functions and parameters are fit to demonstration trajectories using weighted linear regression. Since the NewtonianVAE embeds for proportionality, DMPs can be fit directly to the latent space from demonstration data, allowing for vision-based trajectory control and path following.

\begin{figure*}[!t]
    \centering
    \begin{minipage}{0.63\textwidth}
        \centering
        \includegraphics[width=0.9\textwidth]{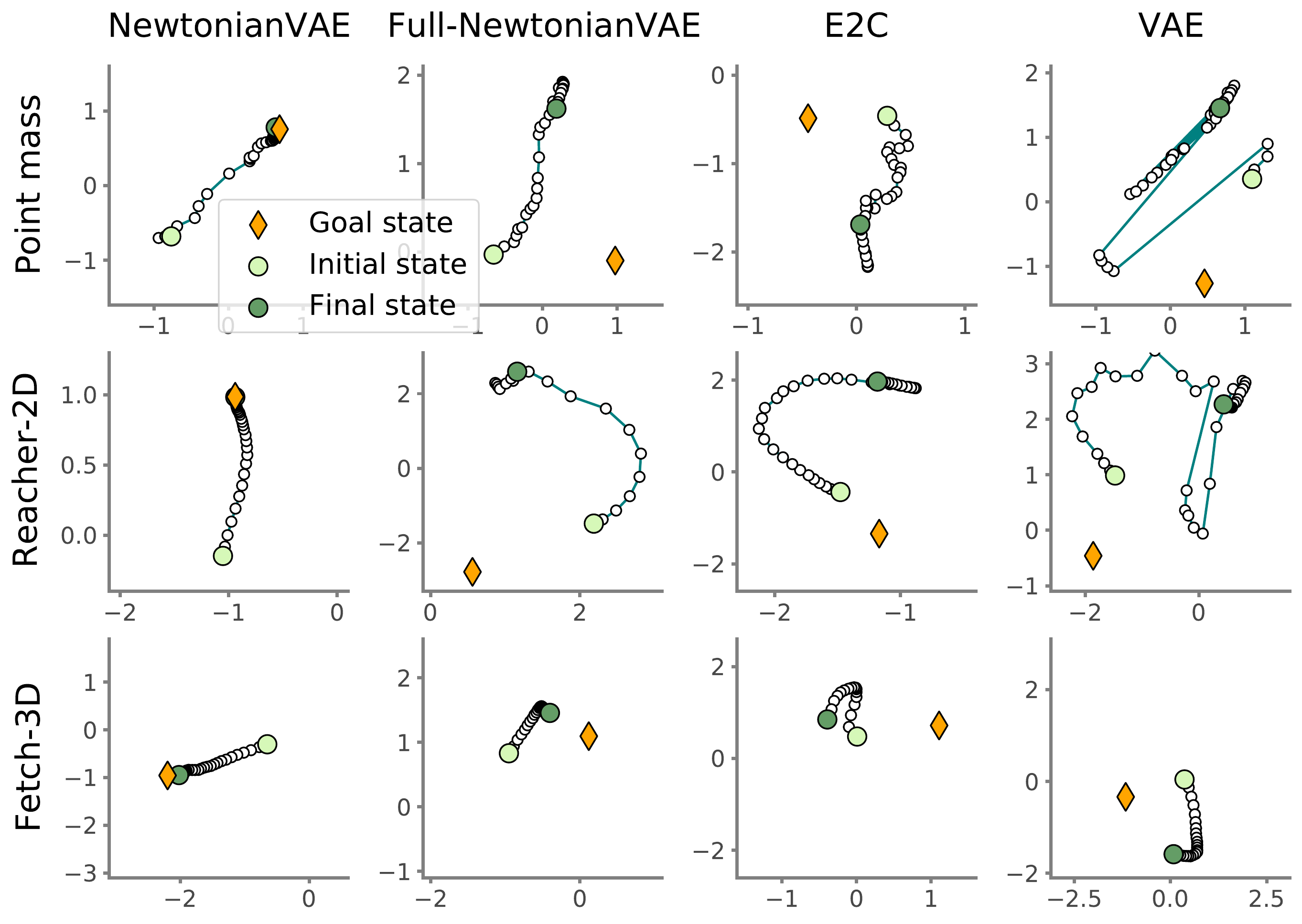}
    \end{minipage}
    \hspace{2mm}
    \begin{minipage}{0.33\textwidth}
       \includegraphics[width=0.9\textwidth]{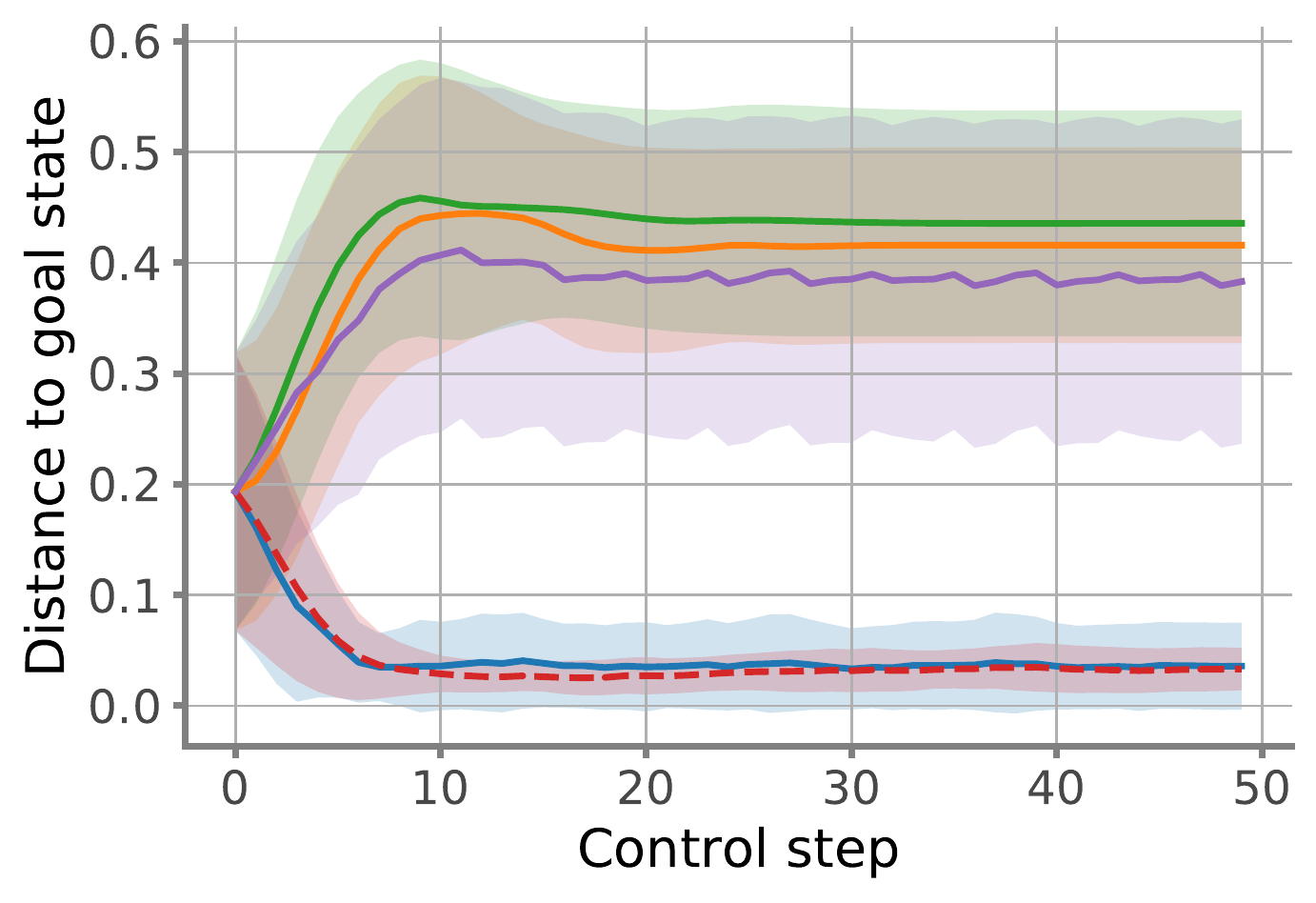}
       \\
       \includegraphics[width=0.9\textwidth]{controlability/convergence_pointmass.pdf}
    \end{minipage}
    \includegraphics[width=0.85\textwidth]{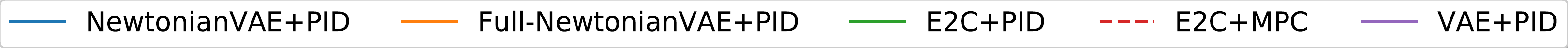}
    \vspace{-1mm}
    \caption{\textbf{Left:} P-control trajectories for point mass, reacher-2D and fetch-3D environments. Plots are in the latent space of Fig. 2. We can see that only NewtonianVAE produces a latent space where a P-controller correctly leads the systems from the initial to goal state. \textbf{Right:} Convergence rates of PID control using various latent embeddings for the point mass (left) and reacher-2D (right) systems, over 50 episodes. We use gain parameters $K_p=8$, $K_i=2$, $K_d=0.5$. For contrast, we show Model Predictive Control (MPC, using CEM planning as per \cite{hafner2018learning}).}
     \label{fig:pc_trajectories}
    \vspace{-3mm}
\end{figure*}

\begin{figure}
  \centering
    \begin{minipage}{0.20\textwidth}
        \centering
        \includegraphics[width=0.99\textwidth]{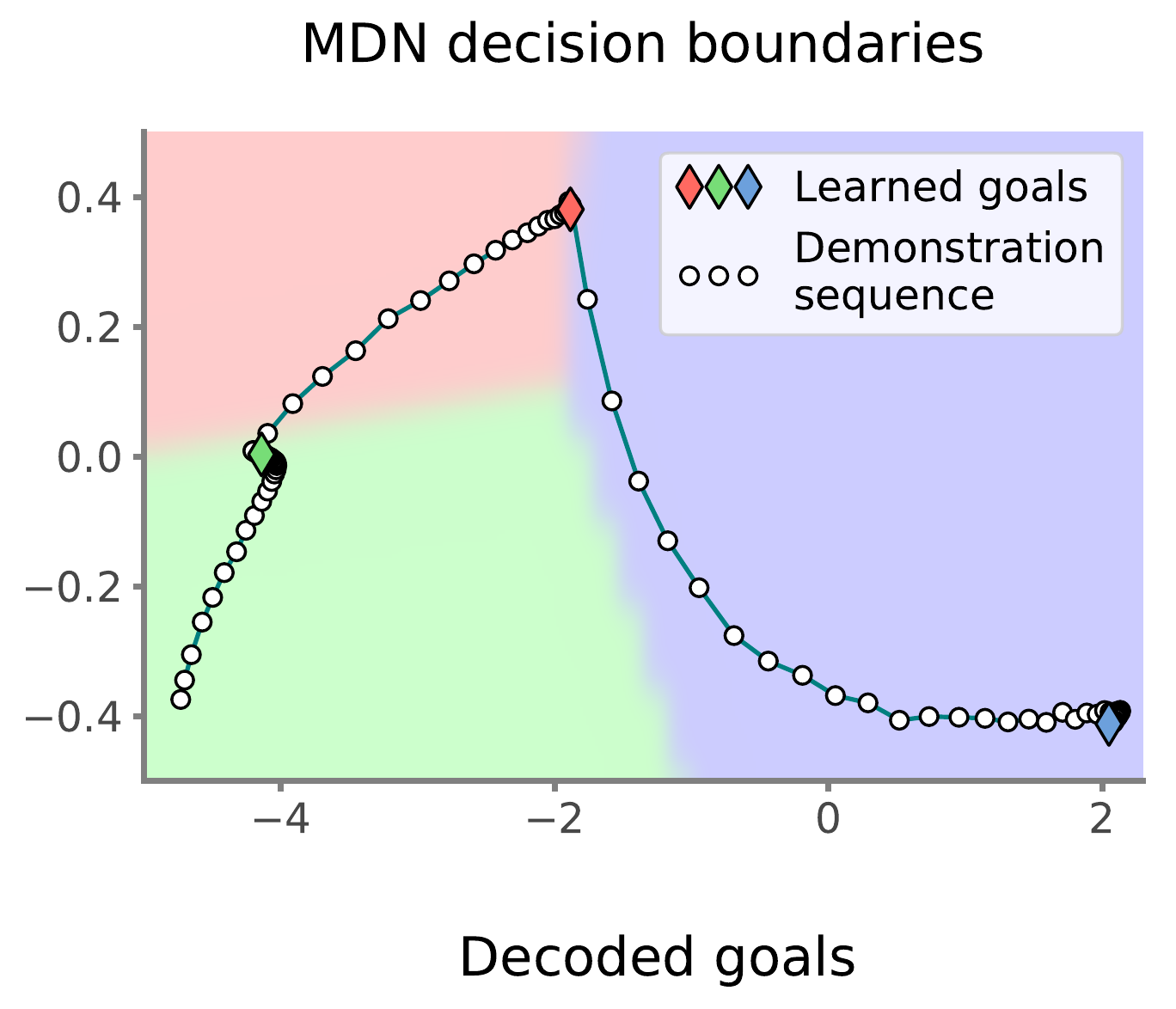}
        \par  \hspace{0.1cm}
        \includegraphics[width=0.80\textwidth]{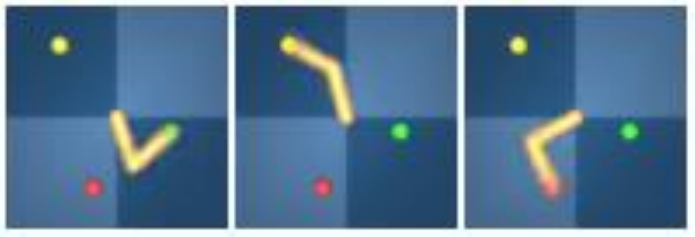}
    \end{minipage}%
    \begin{minipage}{0.27\textwidth}
        \centering
        \includegraphics[width=0.48\textwidth]{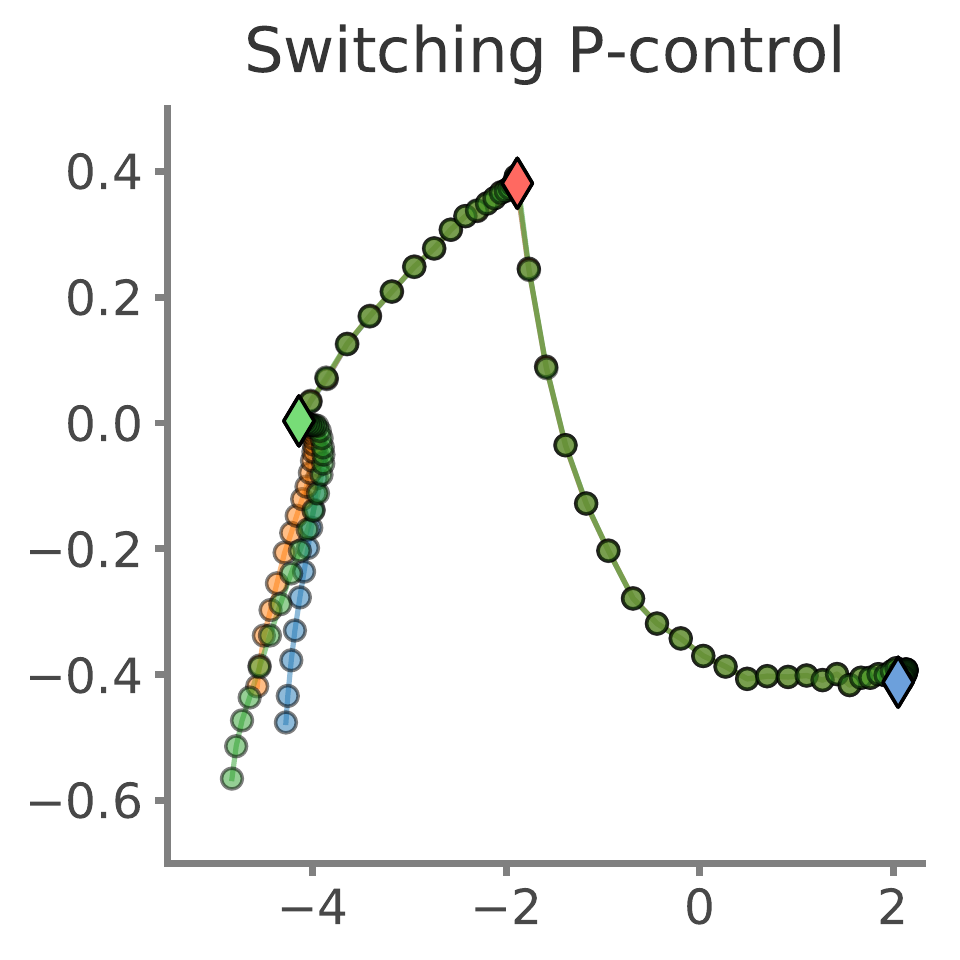}
        \includegraphics[width=0.48\textwidth]{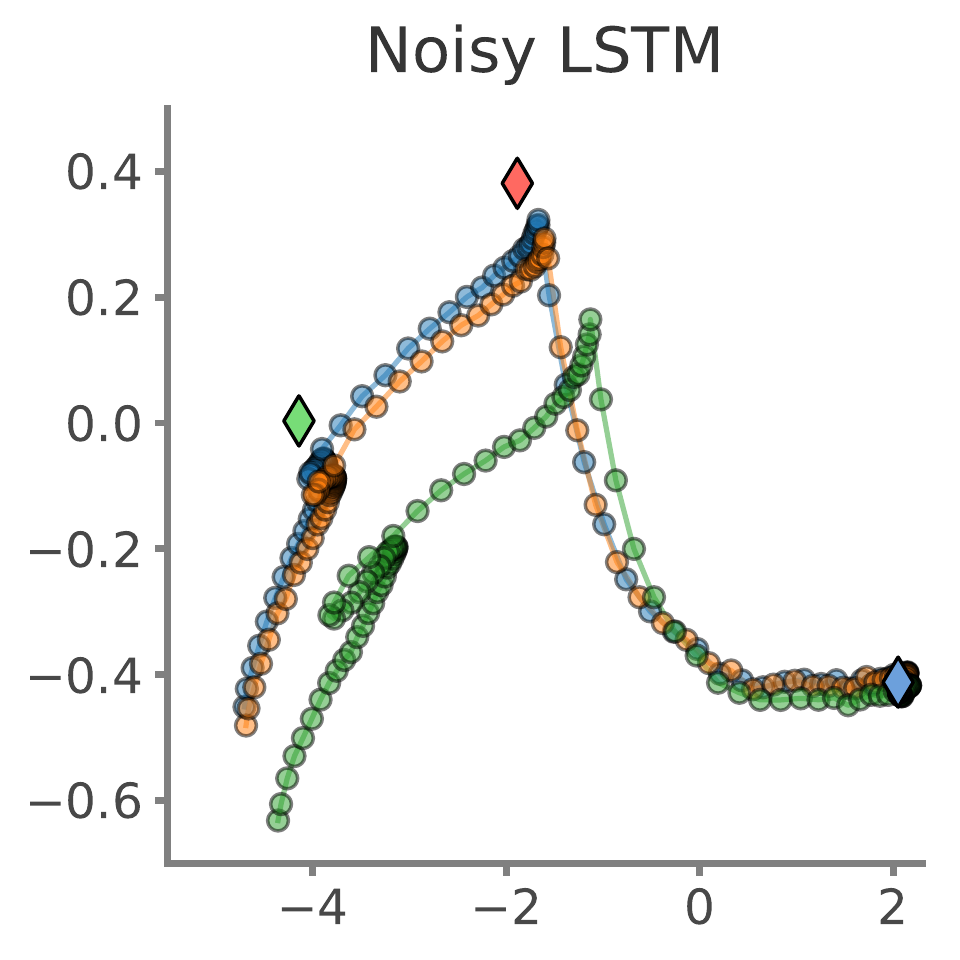}
        \par
        \includegraphics[width=0.48\textwidth]{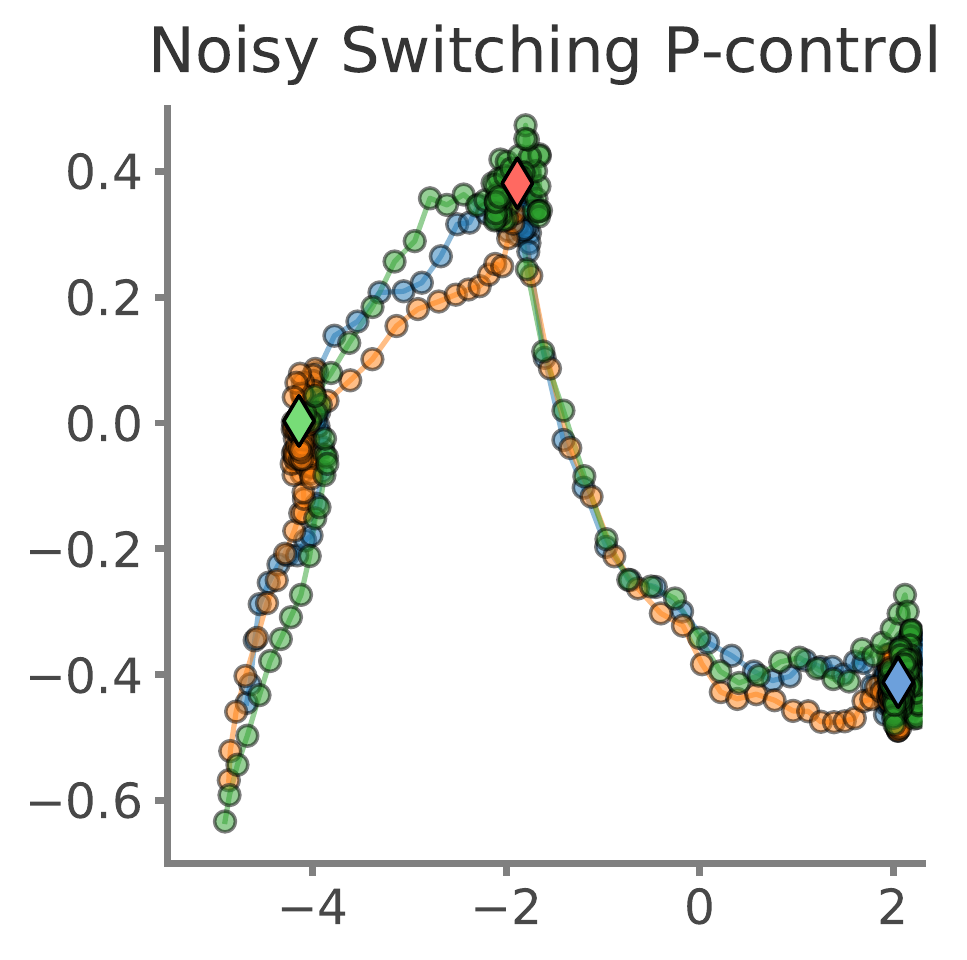}
        \includegraphics[width=0.48\textwidth]{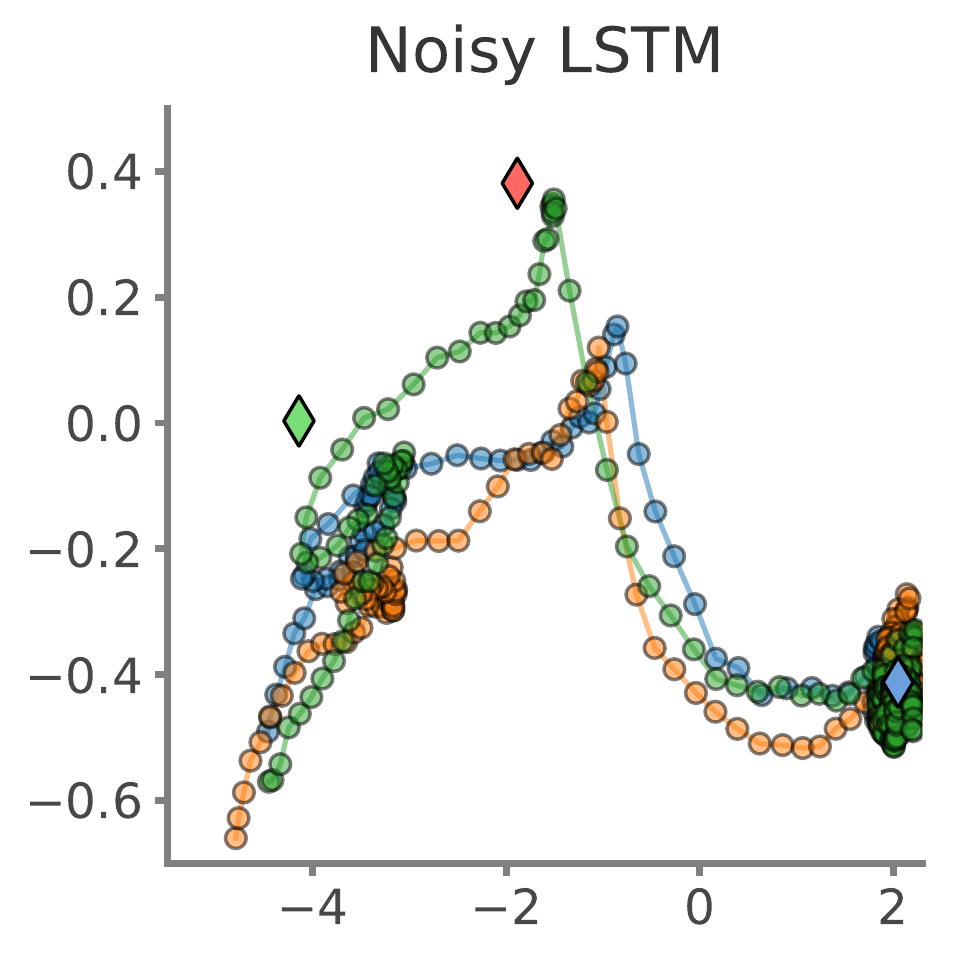}
    \end{minipage}
    \vspace{-2mm}
    \caption{\textbf{Left:} Demonstration sequence and learned mixture of P-controllers (MDN). Each background color and corresponding diamond correspond to a component $\pi_n(\xx)$ and $\xx_n^{goal}$, $\forall n\in\{1,2,3\}$, respectively. \textbf{Right:} Rollouts after imitation learning using switching P-controllers and LSTM policy, with a single demonstration sequence. In the noisy regime each action has an added noise $\mathcal{N}(0,0.25^2)$. All plots are in the NewtonianVAE's latent space.}
    \label{fig:nvae_imitation}
    \vspace{-5mm}
\end{figure}
\section{Experiments}

We validate our model on 3 simulated continuous control environments, to allow for better evaluation and ablations, and on data collected from a real PR2 robot.

\keypoint{Point mass} A simple point mass system adapted from the \verb|PointMass| environment from \verb|dm_control|. The mass is linearly actuated in the 2D plane and its movement bounded by the edges of the frame.

\keypoint{Reacher-2D} A 2D reacher robot adapted from the \verb|Reacher| environment in \verb|dm_control| and inspired by \cite{Kipf2019CompILE:Execution}. We alter the environment so that the robot's middle joint can only bend in one direction, in order to prevent the existence of two possible arm configurations for every end effector position. We also limit the origin joint angle range to $[-160, 160]$ so that the system configuration can be described in polar coordinates by two variables corresponding to the angle of each arm, avoiding a discontinuity in case of full circular motion. 

\keypoint{Fetch-3D} The 3D reacher environment \verb|FetchReachEnv| from OpenAI Gym. We use this to show that our model learns the desirable representations even in visually rich 3D environments of  multi-joint robots with partial occlusions.

To train the models, we generate 1000 random sequences with 100 time-steps for the point mass and reacher-2D systems, and 30 time-steps for the fetch-3D system. More implementation details for each of the environments can be found in Appendix B.

\keypoint{Baseline models} We compare our model to E2C\footnote{DBVF \cite{Karl} and E2C learn similar latent spaces, as both rely on an unstructured conditionally linear dynamical system.} and a static VAE (each frame encoded individually). Additionally, in order to better understand the effect of diagonality and positivity of the transition matrices in \eqref{eq:nvae_trans}, we test Full-NewtonianVAE, where the matrices $A$,$B$,$C$ are unbounded and full rank. 
Architecture and training details can be found in Appendix B. 

\subsection{Visualizing latent spaces and P-controllability}

In this section we compare the latent space and P-controllability properties of the NewtonianVAE and baseline models on the simulated enviroments: point mass, reacher-2D and fetch-3D. 

\keypoint{Comparing latent spaces} We start by visualizing the latent spaces learned by each models on all the environments. Fig. \ref{fig:latent_spaces} shows that only the NewtonianVAE is able to learn a representation corresponding to the natural disentangled coordinates in both environments (e.g. $[x,y]$ in the point mass and $[\theta_1, \theta_2]$ in the reacher-2D), and that these are correctly correlated with ground-truth values, coded in the red-green spectrum. This shows that the structure imposed on the transition matrices in \eqref{eq:nvae_trans} is key to learning correct latent spaces in both Cartesian and polar coordinates. 

\keypoint{P-controllability} Even though the models above produce different latent spaces, most are  well structured and show a clear correlation with the ground truth state (color coded). Although their structure is visually appealing, we are primarily interested in verifying is whether they satisfy P-controllability. To do this, we sample random starting and goal states, and successively apply the control law $\uu_t \propto  (\xx(\II^{goal})-\xx_t(\II_t))$. A space is deemed P-controllable if the system moves to $\xx(\II^{goal})$ in the limit of many time-steps. For reference, we also apply model-predictive control to E2C.    

Convergence curves in the true state space are shown in Fig. \ref{fig:pc_trajectories}, along with example rollouts in the learned latent space (more examples in Appendix C). We can see that only  NewtonianVAE produces P-controllable latent states, as all the remaining models diverge under a P-controller. This highlights the fact that even though the latent spaces learned by the Full-NetwtonianVAE and E2C are seemingly well structured for the point mass system, they fail to provide P-controllability. While these systems can still be stabilised using more complex control schemes such as MPC, this is entirely unnecessary with a P-controllable latent space, where trivial control laws can be applied directly. 


\subsection{MDN goal and boundary visualization}

Having trained a NewtonianVAE on a dataset of random transitions we can use the learned representations to fit the mixture of P-controllers in \eqref{eq:mdn} to the few-shot demonstration sequences. 

\keypoint{Reacher-2D} In this environment there are three colored balls in the scene and the task is reaching the three balls in succession, where the arm's starting location varies across demonstration sequences. We used the true reacher model with a custom controller to generate demonstration images. A full demonstration sequence is shown in Appendix B. For this experiment we use a linear $\bm{\pi}(\xx)$, though a MLP  yields similar results.

After fitting \eqref{eq:mdn} on \textit{a single} demonstration sequence, we visualize the goals $\xx^{goal}$ and the decision boundaries of the switching network $\bm{\pi}(\xx)$ in Fig. \ref{fig:nvae_imitation}(left). The figure shows that  goal states are correctly identified (diamond markers), and that the three sub-task regimes are correctly segmented. Decoding $\xx^{goal}$, confirms that the goals are correctly represented in image space, adding a layer of interpretability to an upstream control policy.

\begin{figure*}[!t]
  \centering
    \includegraphics[width=0.15\textwidth]{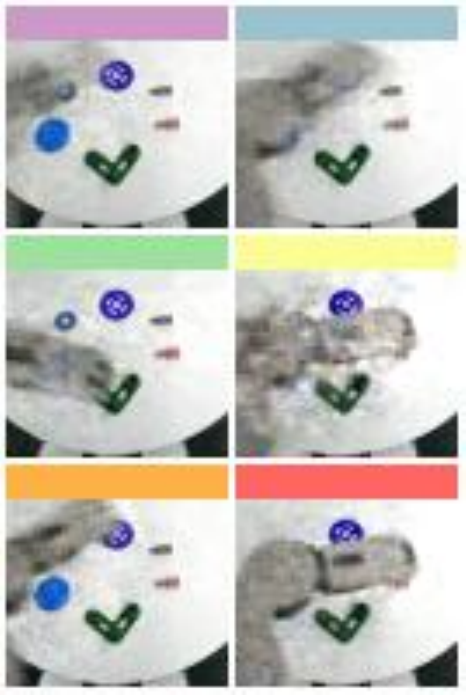}
    \hspace{0.2cm}\vline\hspace{0.2cm} 
    \raisebox{0.1cm}{\includegraphics[width=0.77\textwidth]{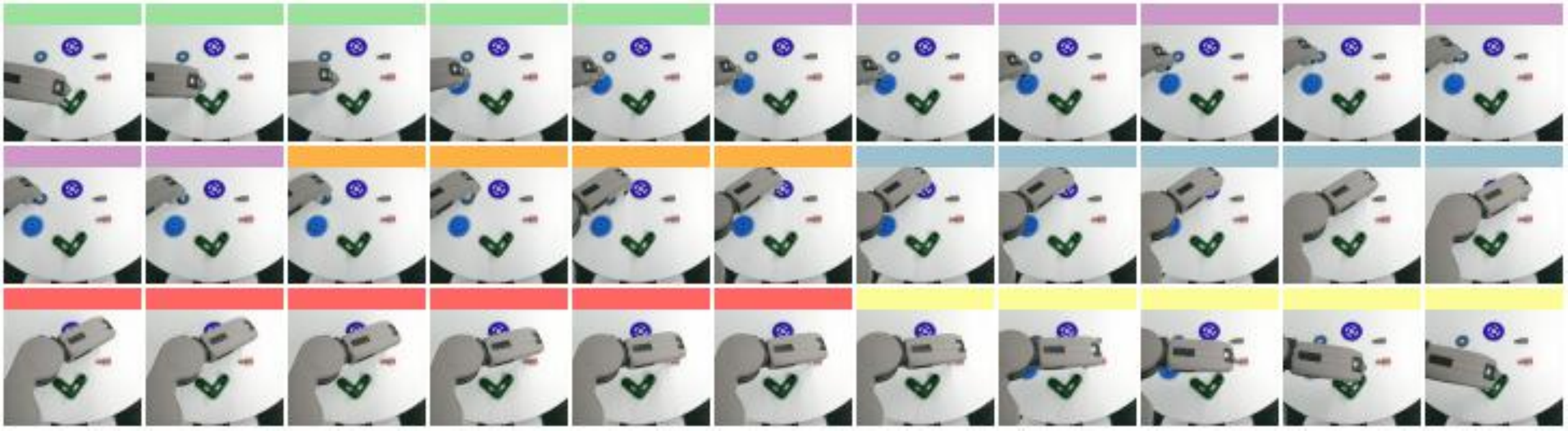}}
    \vspace{-2mm}
    \caption{Decoded goals (left) and sequence segmentation (right) learned for a 6-goal visual trajectory of a PR2 robot. The sequence shows 33 equally spaced frames of a 100-frame demonstration.}
    \label{fig:realreacher}
    \vspace{-3mm}
\end{figure*}

\keypoint{Imitation learning performance} We now compare various imitation learning methods in the simulated task described above. A reward of 1.0 is given when the system reaches a neighborhood of each target (as measured in the true system state), but the targets must be reached in sequence. A more detailed description of the task can be found in Appendix B.  Our method (switching P-controller) uses a finite-state machine inferred from the MDN trained on latent demonstrations (Fig. \ref{fig:nvae_imitation}(left)). We compare it to behaviour cloning with an LSTM with 50 recurrent units, in the NewtonianVAE's latent space, and GAIL \cite{ho2016generative}, a state-of-the-art IRL method trained on ground truth proprioceptive states. Table~\ref{sec:bc} shows the imitation efficiency for increasing numbers of demonstration sequences, with example rollouts shown in Fig.~\ref{fig:nvae_imitation}(right). The results show that goal-driven P-control in a hybrid control policy is significantly more data efficient and robust to noise than a standard behaviour cloning policy. Additionally, switching controllers dramatically outperform GAIL\footnote{Maximum Entropy IRL performed equally poorly, failing to reach a single goal. This is unsurprising, due to the connections between this and adversarial imitation learning \cite{finn2016connection}.}, even though this was trained on 5 times the number of environment interactions used by the NewtonianVAE.

\begin{table*}[!t]
    \centering
    \begin{tabular}{c|cc|cc||c}
       \multirow{2}{*}{\makecell{Demonstration\\sequences}}  & \multicolumn{2}{c|}{Switching P-controller} & \multicolumn{2}{c||}{LSTM} & \multirow{2}{*}{\makecell{GAIL from\\proprioception}} \\ 
        & Clean & Noisy & Clean & Noisy & \\ \hline
        1  & $3.0\pm0.0$ & $2.17\pm0.32$ & $0.81\pm0.35$ & $0.27\pm0.20$ & ---  \\
        10 & $3.0\pm0.0$ & $2.01\pm0.34$ & $3.00\pm0.00$ & $1.42\pm0.34$ & --- \\
        100 & $3.0\pm0.0$ & $2.06\pm0.30$ & $3.00\pm0.00$ & $1.23\pm0.30$ & 0.62 \\
    \end{tabular}
    \vspace{-2mm}
    \caption{Efficiency of imitation learning methods for vision-based sequential multi-task control. Metric: Environment Reward (max = 3.0). The NewtonianVAE is used to encode the frames. `Noisy': Added action noise $\mathcal{N}(0,0.25^2)$ during the rollouts. Error ranges: 95\% confidence interval across 100 rollouts. GAIL is trained for 5000 episodes.}
    \label{sec:bc}
    \vspace{-5mm}
\end{table*}

\keypoint{Real multi-object reacher} 
We now apply our model to real robot data. Here, we record a 7-DoF PR2 robot arm that moves between 6 objects in succession in a hexagon pattern. A full sequence comprises approximately 100 frames. We use 636 frames to train the NewtonianVAE and an additional 100 held-out frames to train the MDN. Further model and dataset details can be found in Appendix B. 

Fig. \ref{fig:realreacher} shows the image representations of the learned goals (left) and the mode $\bm{\pi}(\xx)$ that is active for each frame in the demonstration sequence (right). We can see that the six goals are correctly identified by the MDN, and that  segmentations are correct in the sense that a frame is assigned to the learned goal to which the robot is moving at that time step. Note that the model is able to recover correct goals and segmentations even though not all of the joints are visible in every frame.

\subsection{Fitting DMPs for path following in latent space}
We show how the NewtonianVAE can be used to enable a robot to learn a vision-driven controller to follow a demonstration trajectory, using the fetch-3D environment. To this end, we draw a 'G'-shaped trajectory in the first 2 dimensions of the latent space and fit a DMP. The DMP runs in 100 time-steps, spanning 4 seconds of execution, where we feed the acceleration output by the DMP as the action to the environment, and the new state and velocity is inferred by the NewtonianVAE.

\begin{figure}[h]
    \centering
    \includegraphics[width=0.20\textwidth]{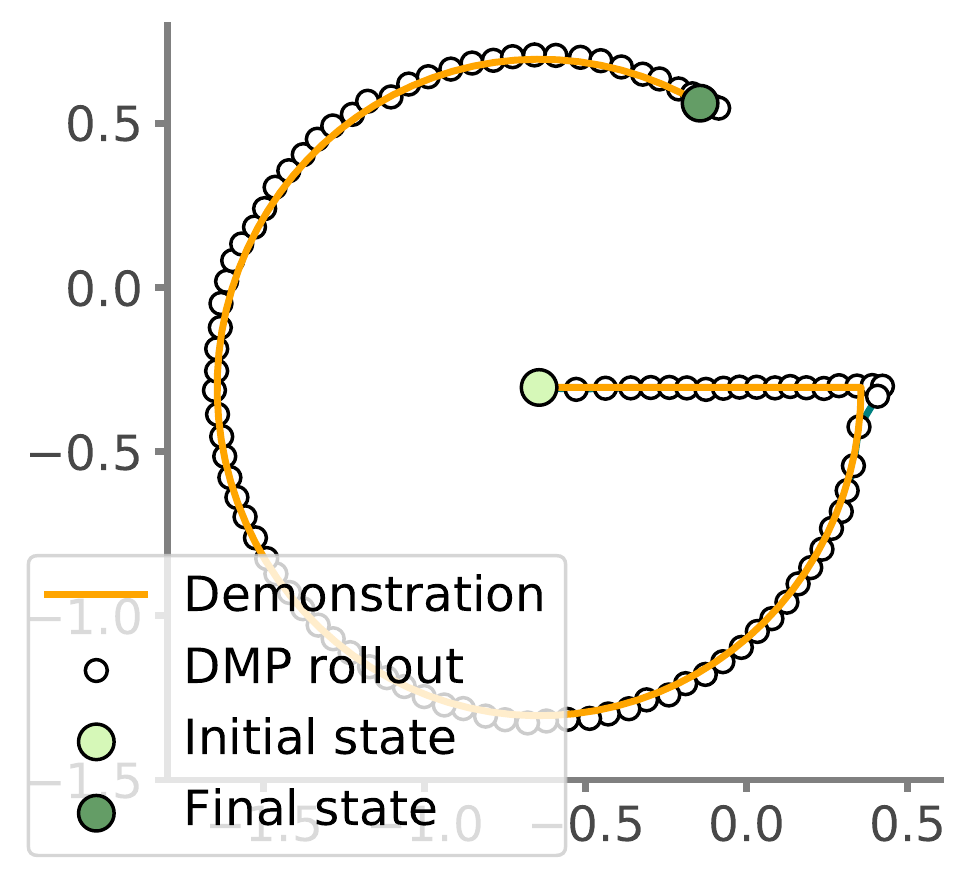}
    \includegraphics[width=0.26\textwidth]{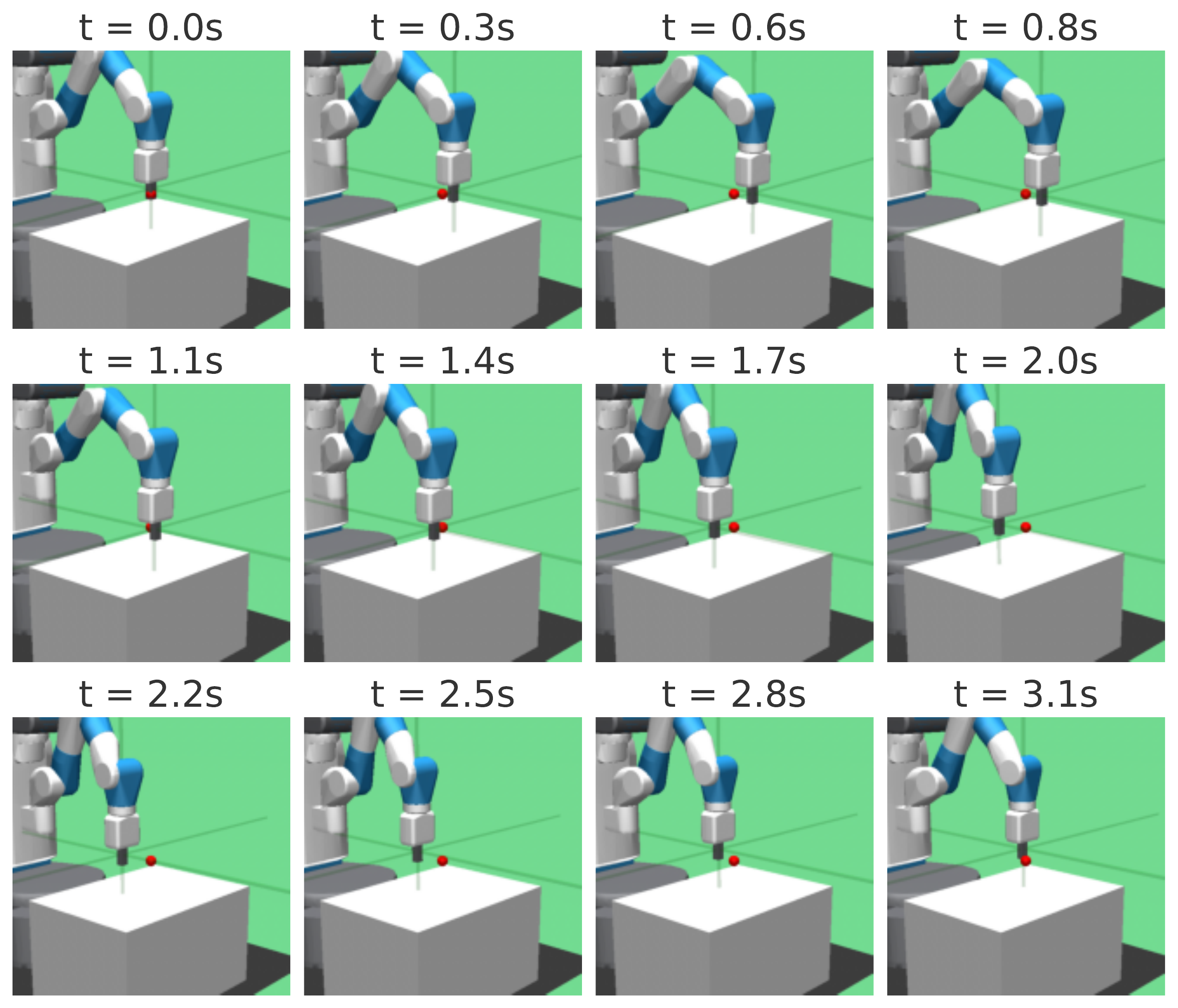}
    \caption{\textbf{Left:} Overhead view of demonstration and trajectory produced by the DMP in the fetch-3D environment. The  first 2 dimensions of the NewtonianVAE's latent space are shown. \textbf{Right:} Frames seen by the NewtonianVAE during this rollout.}    
    \label{fig:dmp_traj}
    \vspace{-4mm}
\end{figure}

Fig. \ref{fig:dmp_traj} shows that the robot correctly follows the demonstration trajectory, showing that the latent space induced by the NewtonianVAE enables path following using a DMP just by virtue of its P-controlability property, without needing to be explicitly trained to perform well under a DMP, as done by \cite{chen2016dynamic}. 

\section{Discussion}
\keypoint{Limitations and Future Work}
This work assumes that underlying systems are proportional controllable, and follow Newtonian dynamics. Moreover, it should be noted that vision-based torque control of high dimensional robot manipulators requires high speed vision. However, in our opinion, the most notable limitation is the fact that the imitation learning model only learns a fixed set of goals. Ideally, the agent would learn a semantic goal, which would represent a command "fetch the yellow ball", for a variable position of the yellow ball and not a fixed  state. However, this would require demonstration data with substantially more variety than considered here. We have also avoided multi-modal demonstrations for simplicity, though we believe it would be of interest to integrate our method with approaches like InfoGAIL \cite{Li2017InfoGAIL:Demonstrations}.

\keypoint{Conclusion}
We introduced NewtonianVAE, a structured latent dynamics model designed to allow P-controllability from pixels. Results show that this structured latent space allows for trivial, robust control in the presence of noise and dramatically simplifies and improves imitation learning, which can be framed either as a switching goal-inference or as a path following problem in the latent space.
Additionally, our model provides visually interpretable goal discovery and task segmentation under both simulated and real environments, without any labelled or proprioception data.

{\small
\bibliographystyle{abbrvnat}
\bibliography{imitation_learning_refs.bib,latent_dynamics_refs.bib,references.bib}
}

\newpage

\appendix

\section{ELBO derivation}
\label{sec:elbo}

We want to maximize the sequence marginal likelihood:
\begin{equation}
    p(\II_{1:T}|\uu_{1:T}) = \int p(\II_{1:T}|\xx_{1:T},\uu_{1:T})p(\xx_{1:T}|\uu_{1:T}) \, \text{d}\xx_{1:T}.
    \label{eq:appendix_marginal}
\end{equation}
We factorize the above terms as follows:
\begin{align}
    p(\II_{1:T}&|\xx_{1:T},\uu_{1:T}) = \prod_t p(\II_t|\xx_{t-1},\uu_{t-1}) \nonumber \\
    &= \prod_t \int p(\II_t|\hat{\xx}_t) p(\hat{\xx}_t|\xx_{t-1},\uu_{t-1}; \vv_{t-1}) \, \text{d} \hat{\xx}_t \\
    p(\xx_{1:T}&|\uu_{1:T}) = \prod_t p(\xx_t|\xx_{t-1},\uu_{t-1}; \vv_{t-1}),
\end{align}
where $\vv_t=(\xx_t-\xx_{t-1})/\Delta t$. Hence, $p(\xx_t|\xx_{t-1},\uu_{t-1}; \vv_{t-1})$ depends on $\xx_{t-2}$ through $\vv_{t-1}$, but we will simply use $p(\xx_t|\xx_{t-1},\uu_{t-1}; \vv_{t-1}) \equiv p(\xx_t|\xx_{t-1},\uu_{t-1})$ for ease of readability. We use an approximate posterior factorized as:
\begin{equation}
    q(\xx_{1:T}|\II_{1:T}) = \prod_t q(\xx_t|\II_t)
\end{equation}

Using Jensen's inequality we can write \eqref{eq:appendix_marginal} as:

\begingroup
\allowdisplaybreaks
\begin{align}
    &\log p(\II_{1:T}|\uu_{1:T}) = \\
    &= \log \Bigl( \int  \frac{\prod_t q(\xx_t|\II_t)}{\prod_t q(\xx_t|\II_t)} \prod_t \int p(\II_t|\hat{\xx}_t) p(\hat{\xx}_t|\xx_{t-1},\uu_{t-1}) \, \text{d} \hat{\xx}_t \nonumber \\
    & \hspace{6mm} \prod_t p(\xx_t|\xx_{t-1},\uu_{t-1})  \text{d}\xx_{1:T} \Bigr) \\
    &\geq \int \prod_t q(\xx_t|\II_t) \Bigg( \sum_t \log \left[ \int p(\II_t|\hat{\xx}_t) p(\hat{\xx}_t|\xx_{t-1},\uu_{t-1}) \, \text{d} \hat{\xx}_t \right] \nonumber  \\
    &  \hspace{6mm} + \sum_t \log \frac{p(\xx_t|\xx_{t-1},\uu_{t-1})}{q(\xx_t|\II_t)} \Bigg) \text{d}\xx_{1:T} \\
    &= \sum_t \int  q(\xx_{t-1}|\II_{t-1}) q(\xx_{t-2}|\II_{t-2}) \nonumber \\ 
    & \hspace{6mm} \Bigg( \log \left[ \int p(\II_t|\hat{\xx}_t) p(\hat{\xx}_t|\xx_{t-1},\uu_{t-1}) \, \text{d} \hat{\xx}_t \right]  \text{d}\xx_{t-1} \nonumber  \\
    & \hspace{6mm}+ \text{KL}\left(q(\xx_t|\II_t)\|p(\xx_t|\xx_{t-1},\uu_{t-1}) \right) \Bigg) \\
    &\geq  \sum_t  \mathbb{E}_{q(\xx_{t-1} | \II_{t-1}) q(\xx_{t-2}|\II_{t-2})} \Big( \mathbb{E}_{p(\hat{\xx}_t|\xx_{t-1},\uu_{t-1})} \log p(\II_t|\hat{\xx}_t) \nonumber \\
    & \hspace{6mm} + \text{KL}\left(q(\xx_t|\II_t)\|p(\xx_t|\xx_{t-1},\uu_{t-1})\right) \Big) \label{eq:appendix_elbo}
\end{align}
\endgroup

\section{Full experimental details}
\label{sec:details}

\subsection{Architectures}

All models used the encoder and decoder from \citet{ha2018world}, except for the point mass environment, where we use a spatial broadcast decoder \cite{Watters2019SpatialVAEs}. All temporal models were trained using 2-step ahead prediction in the ELBO (instead of single step), which is a straightforward extension of \eqref{eq:nvae_elbo}, as done in latent overshooting \cite{hafner2018learning}. All experiments use $64\times64$ RGB frames as input to the encoder. To compute the transition matrices as a function of the state we use a fully connected network with 2 hidden layers with 16 units and ReLU activation, with the appropriate input and output dimensionality. In the NewtonianVAE variants, $\Delta t$ was set to the known environment time step. All models were trained using Adam \cite{Kingma} with a learning rate of $3\cdot 10^{-4}$ and batch size 1 (a single sequence per batch) for 300 epochs. In the point mass experiments we found it useful to anneal the KL term in the ELBO, starting with a value of 0.001 and increasing it linearly to 1.0 between epochs 30 and 60.

\subsection{Simulated point mass environment}

The point mass environment is adapted from the \verb|PointMass| environment from the \verb|dm_control| library. The mass is linearly actuated in the 2D plane and its movement bounded by the edges of the frame. The simulator uses a time-step $\Delta t=0.5$ and the $[x,y]$ forces are in the range $[-1,1]^2$.

\subsection{Simulated reacher environment}

\begin{figure*}[!h]
    \centering
    \includegraphics[width=0.7\textwidth]{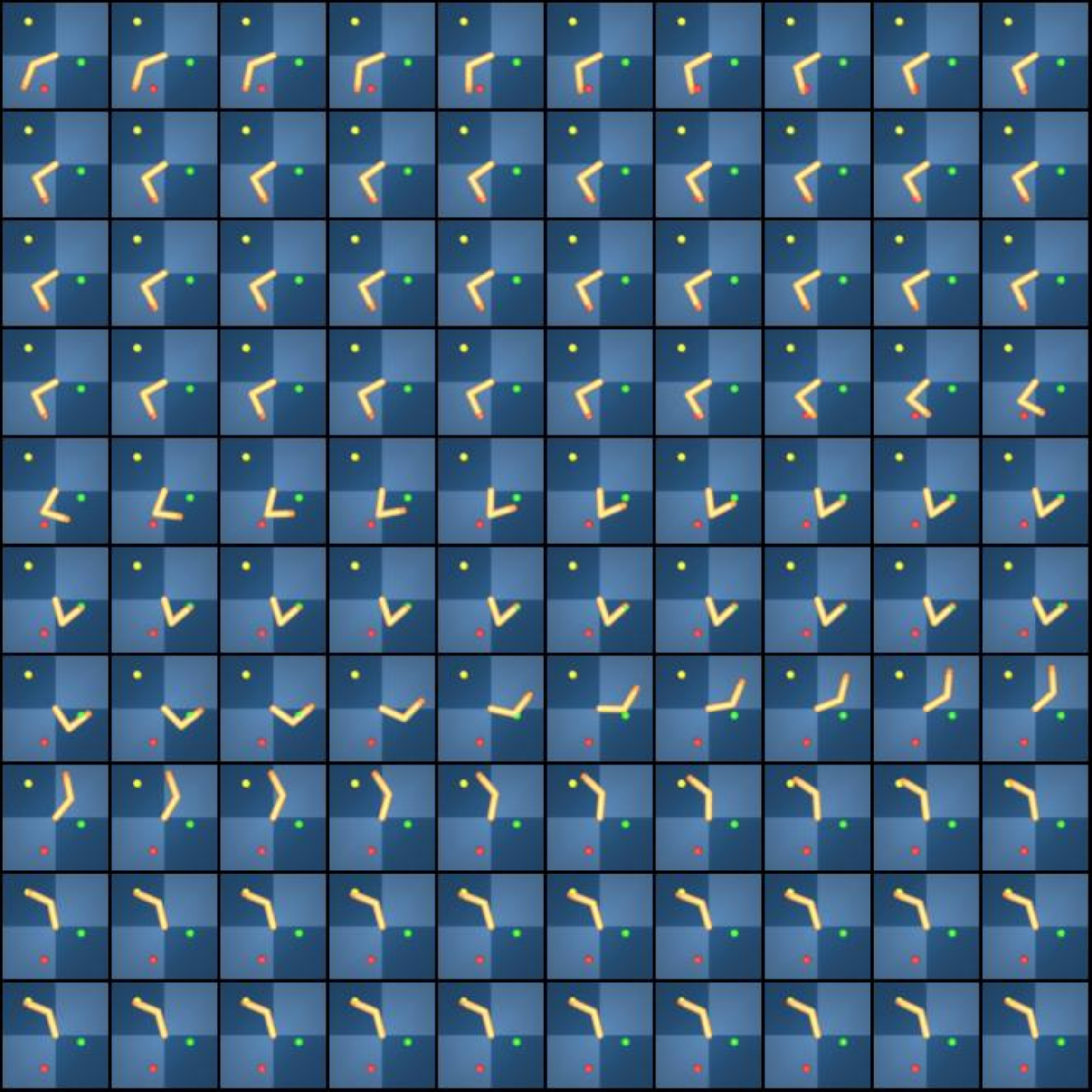}
    \caption{Full demonstration sequence for simulated reacher (progression left to right, top to bottom).}
    \label{fig:appendix_demo}
\end{figure*}

 The reacher-2D environment is a adapted from the \verb|Reacher| environment and inspired by the simulated reacher task in \citet{Kipf2019CompILE:Execution}. We limit the rotation of the shoulder joint to the $[-160, 160]$ range, and the wrist joint to $[0, 160]$. The simulator uses a time-step of $\Delta t=0.1$ and the torques are in the range $[-1,1]$. When generating random rollouts we sample shoulder and wrist angles in the whole range, and when generating demonstrations these angles are sampling according to \verb|0.5+(np.random.rand()-0.5)| and \verb|-np.pi+0.3+np.random.rand()*0.5| (in radians), respectively. A full 100-step demonstrations sequence is shown in Fig. \ref{fig:appendix_demo}.
 
 To evaluate the trained control policies in the simulator, we compute a sparse reward as follows. When the distance between the end affector and the initial target is lower than 0.015 and the joint velocity is lower than 0.2, the agent earns a reward of 1. The targets must be reached in sequence, i.e., if the agent goes straight for the second target without stopping at the first target, the reward is still 0. The optimal agent will thus have a maximum reward of 3. We use 120- and 220-step rollouts when evaluating the noiseless and noisy settings, respectively.

\label{sec:appendix_reacher}

\subsection{PR2 robot arm}
\label{sec:appendix_pr2}

Real robot experiments were conducted using the left arm of a PR2 robot, with images recorded using a downward facing Kinect 2 camera mounted on the PR2 head. Arm motion demonstrations were obtained by pre-programming the robot to move to various objects in the scene using the MoveIt! motion planning library in the robot operating system (ROS). The robot arm is actuated using 8 torque commands (7 for the joints in the robot arm and one for the robot torso height), which were recorded alongside images.

After preprocessing the images (rescaling to $64\times64$ and cropping to the region of interest), we are left with 836 frames, which we split into 636 training, 100 validation, and 100 testing frames. Training used a batch size of 20 frames.

Due to the very small amount of training data available, we had to impose further constraints on the model to allow for correct learning. Firstly, the transition matrices were set to $A=0$, $B=0$, $C=1$. Secondly, we added an additional regularization term to the latent space, $KL(q(\xx|\II)\| \mathcal{N}(0,1))$, to improve visualization of the goals (though this term was not necessary for obtaining correct sequence segmentations). Finally, we added a batch-wise entropy term in $\bm{\pi}(\xx)$ to encourage the use of all modes, as proposed by \cite{burke2019hybrid}:

\begin{equation}
    \mathcal{L}^{\text{ENT}} = - \frac{1}{J} \sum_{j=1}^J \log \left(\frac{1}{T}\sum_{t=1}^T \pi_{j,t} \right).
\end{equation}

\section{P-control trajectories}
\label{sec:trajectories}

\begin{figure*}[!h]
    \centering
    \includegraphics[width=0.95\textwidth]{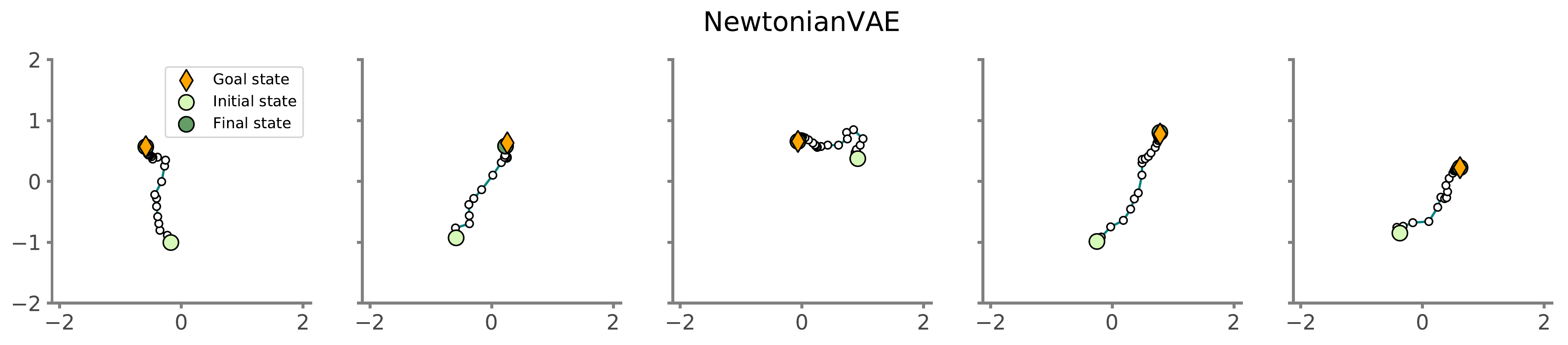}
    \par \vspace{0.4cm}
    \includegraphics[width=0.95\textwidth]{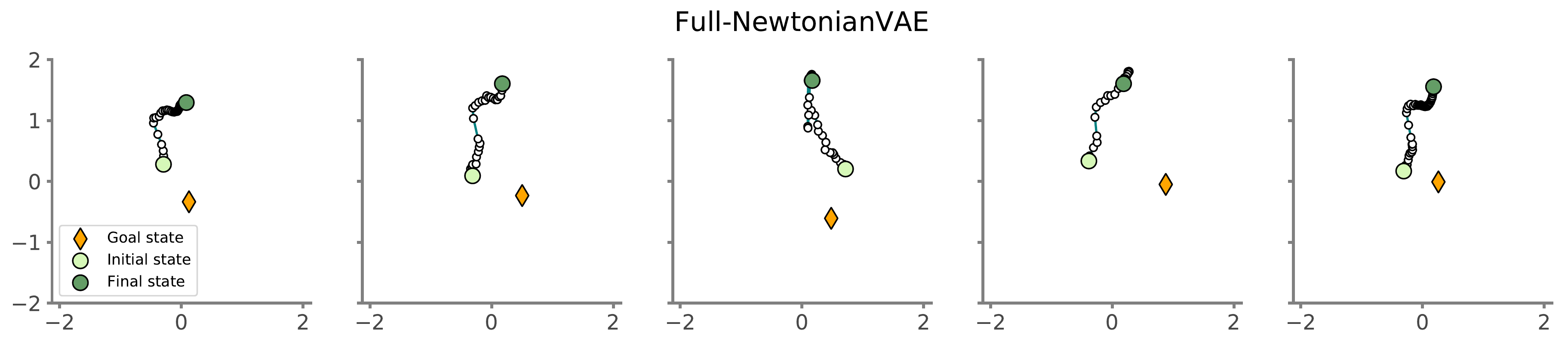}
    \par \vspace{0.4cm}
    \includegraphics[width=0.95\textwidth]{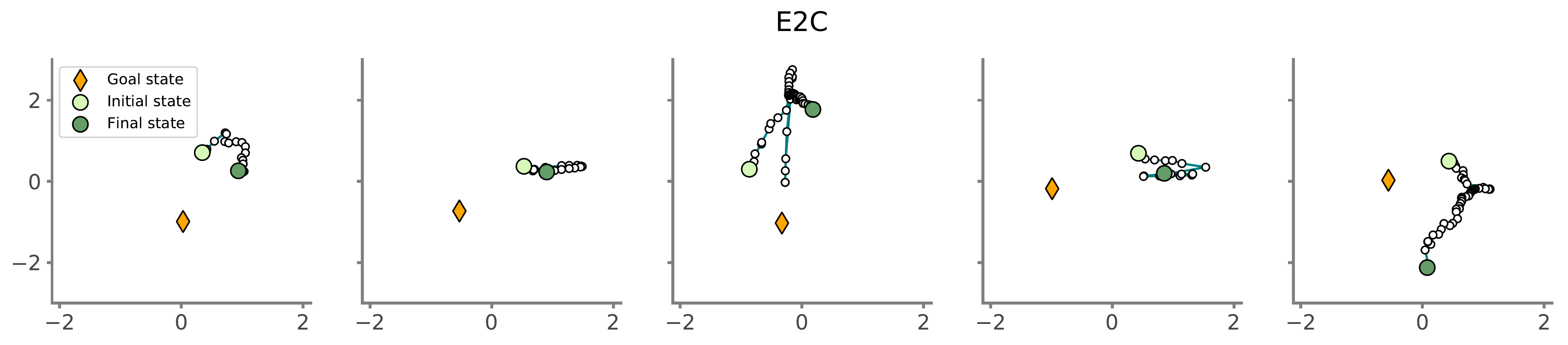}
    \par \vspace{0.4cm}
    \includegraphics[width=0.95\textwidth]{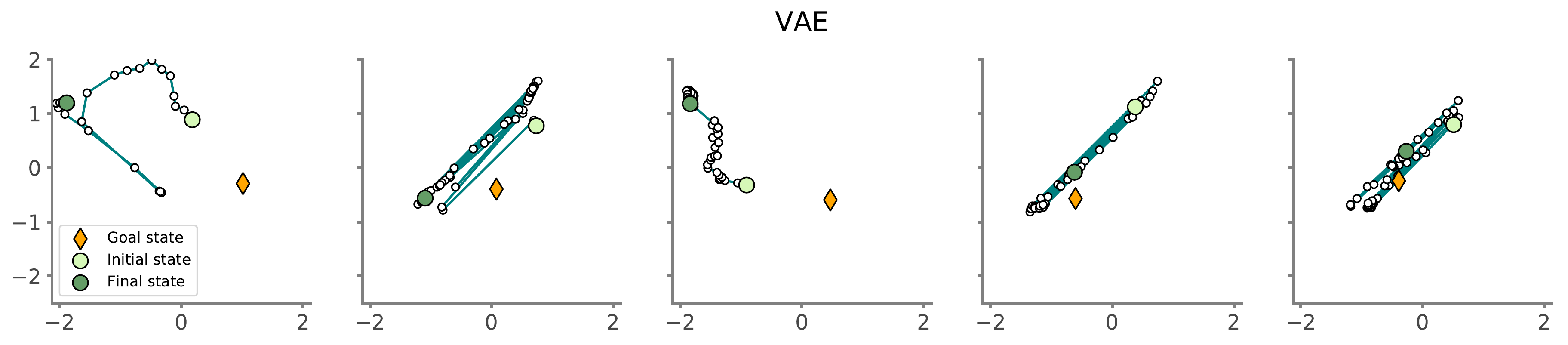}
    \caption{P-controllability in point mass system.}
    \label{fig:appendix_pointmass_traj}
\end{figure*}

\begin{figure*}[!h]
    \centering
    \includegraphics[width=0.95\textwidth]{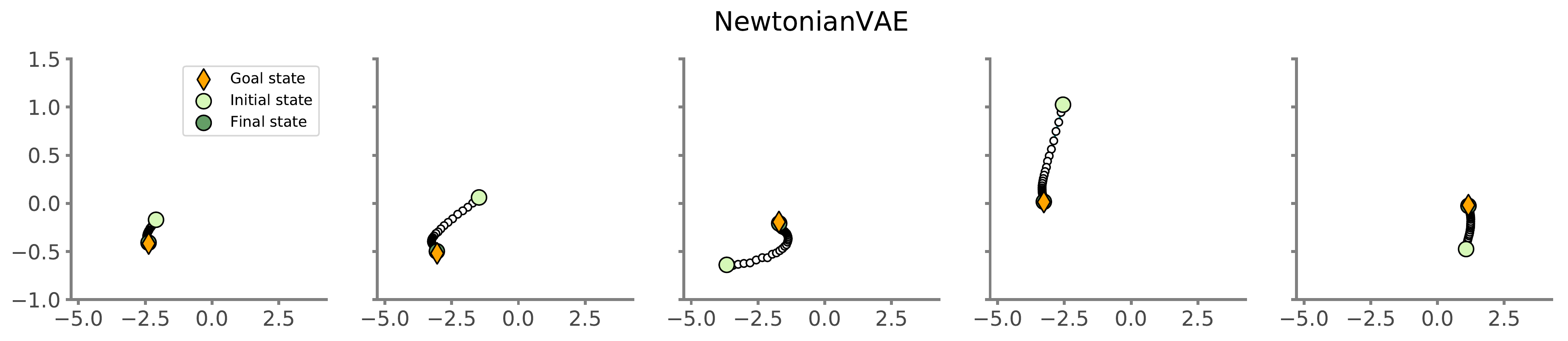}
    \par \vspace{0.4cm}
    \includegraphics[width=0.95\textwidth]{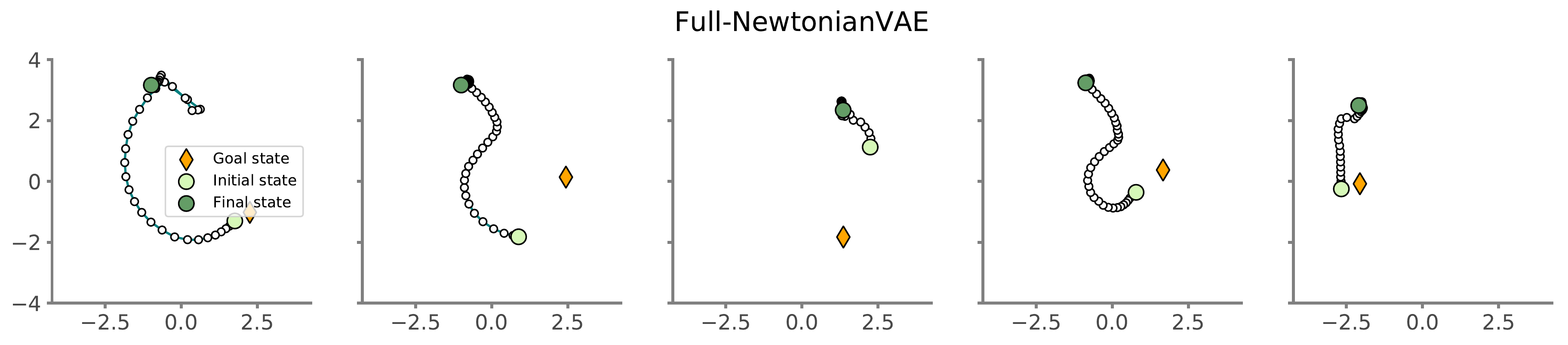}
    \par \vspace{0.4cm}
    \includegraphics[width=0.95\textwidth]{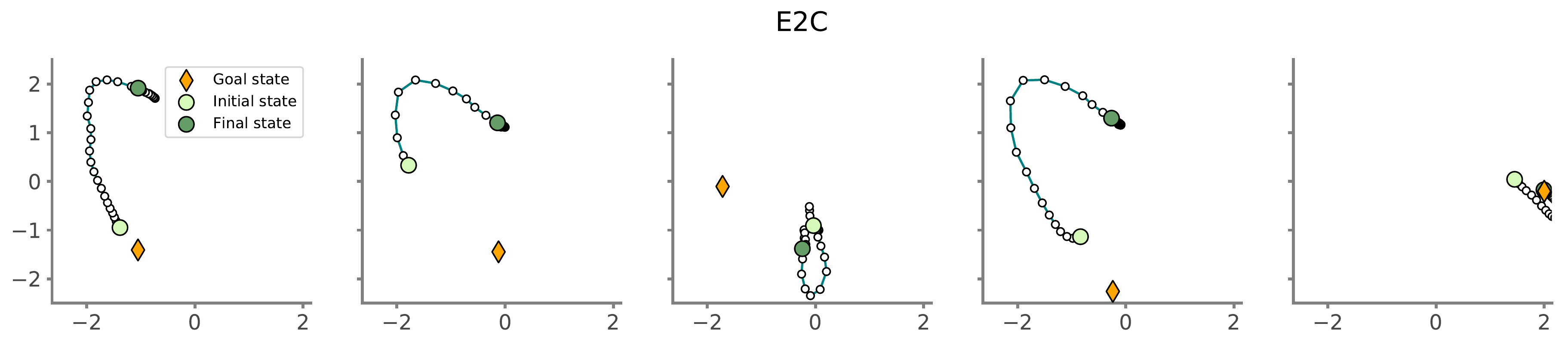}
    \par \vspace{0.4cm}
    \includegraphics[width=0.95\textwidth]{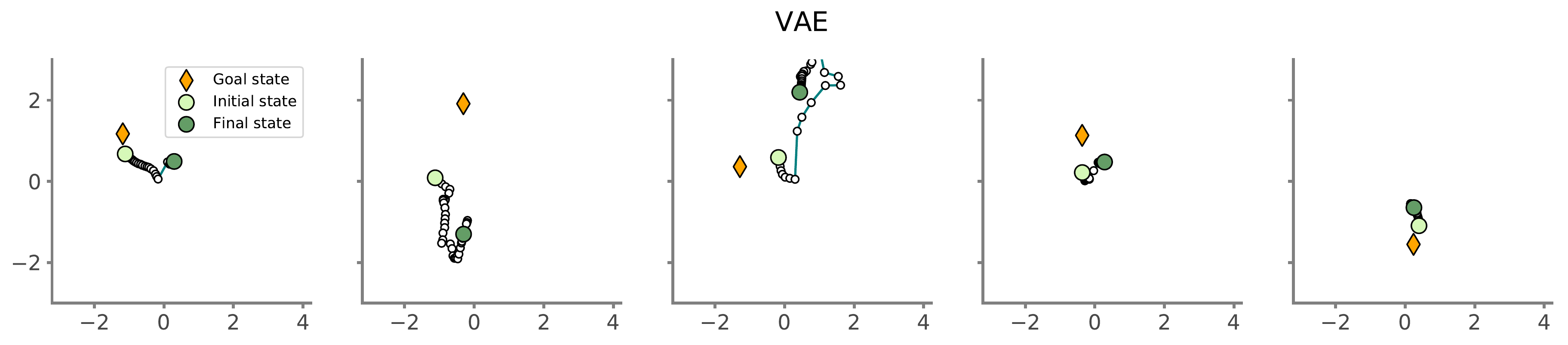}
    \caption{P-controllability in reacher system.}
    \label{fig:appendix_reacher_traj}
\end{figure*}

\begin{figure*}[!h]
    \centering
    \includegraphics[width=0.95\textwidth]{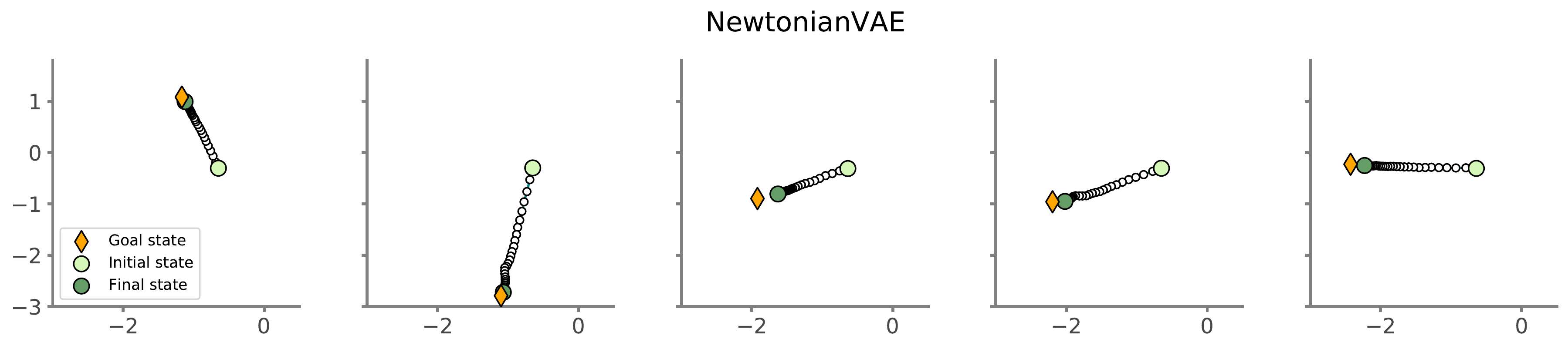}
    \par \vspace{0.4cm}
    \includegraphics[width=0.95\textwidth]{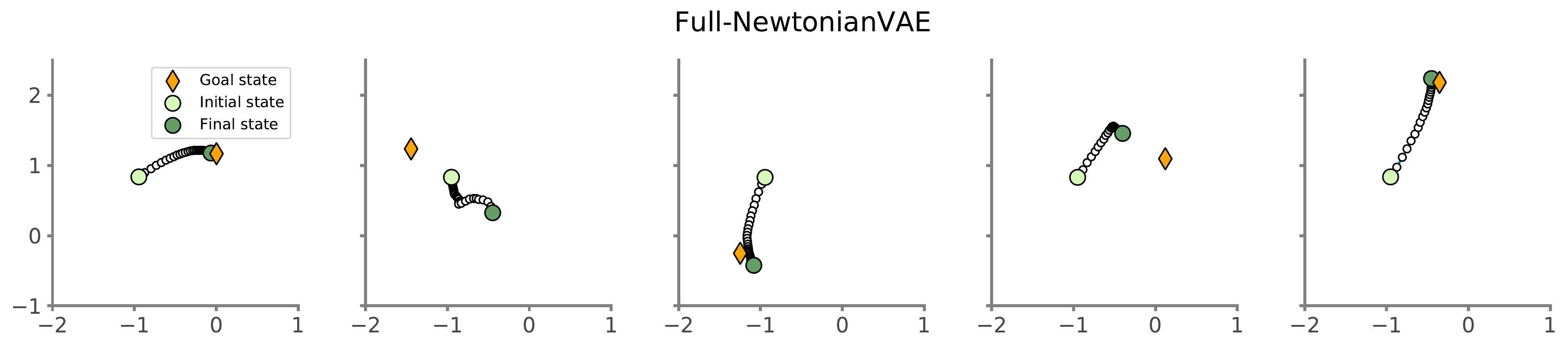}
    \par \vspace{0.4cm}
    \includegraphics[width=0.95\textwidth]{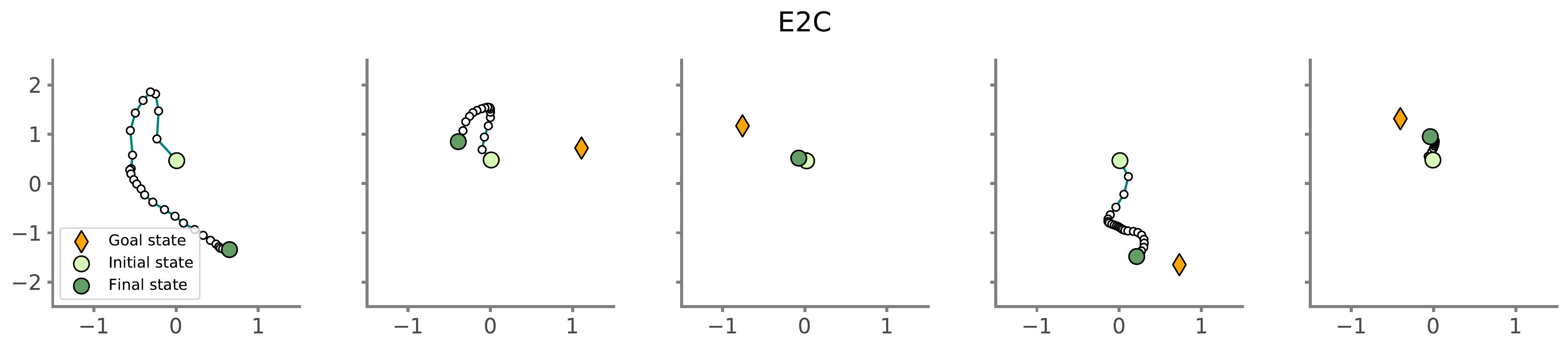}
    \par \vspace{0.4cm}
    \includegraphics[width=0.95\textwidth]{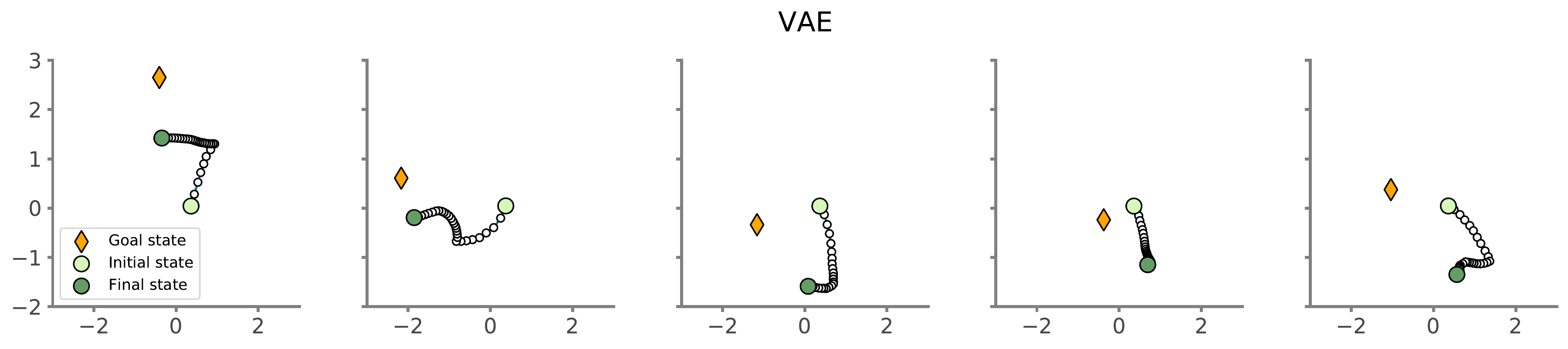}
    \caption{P-controllability in reacher system.}
    \label{fig:appendix_fetch_traj}
\end{figure*}

Additional P-control trajectories for the point mass, reacher-2D, and fetch-3D systems are shown in Figs. \ref{fig:appendix_pointmass_traj},  \ref{fig:appendix_reacher_traj} and \ref{fig:appendix_fetch_traj}, respectively.

\end{document}